\documentclass[runningheads]{llncs}
\usepackage{graphicx}
\usepackage{tikz}
\usepackage{comment}
\usepackage{amsmath,amssymb} % define this before the line numbering.
\usepackage{color}
\usepackage{orcidlink}
\usepackage{booktabs}
\usepackage{algorithm}
\usepackage{algorithmic}
\usepackage{bm}
\usepackage{multirow}

\usepackage[accsupp]{axessibility}  % Improves PDF readability for those with disabilities.

\begin{document}
% \renewcommand\thelinenumber{\color[rgb]{0.2,0.5,0.8}\normalfont\sffamily\scriptsize\arabic{linenumber}\color[rgb]{0,0,0}}
% \renewcommand\makeLineNumber {\hss\thelinenumber\ \hspace{6mm} \rlap{\hskip\textwidth\ \hspace{6.5mm}\thelinenumber}}
% \linenumbers
\pagestyle{headings}
\mainmatter
\def\ECCVSubNumber{4032}  % Insert your submission number here

\title{DeltaGAN: Towards Diverse Few-shot Image Generation with Sample-Specific Delta} % Replace with your title

% INITIAL SUBMISSION 
\begin{comment}
\titlerunning{ECCV-22 4032 \ECCVSubNumber} 
\authorrunning{ECCV-22 4032 \ECCVSubNumber} 
\author{Anonymous ECCV submission}
\institute{4032 \ECCVSubNumber}
\end{comment}
%******************

% CAMERA READY SUBMISSION
%\begin{comment}
\titlerunning{DeltaGAN}
% If the paper title is too long for the running head, you can set
% an abbreviated paper title here
%
\author{Yan Hong\orcidlink{0000-0001-6401-0812}\index{Yan Hong} \and
Li Niu\thanks{Corresponding author.}  \orcidlink{0000-0003-1970-8634} \and
Jianfu Zhang\orcidlink{0000-0002-2673-5860} \and
Liqing Zhang\orcidlink{0000-0001-7597-8503}
}

\authorrunning{Y. Hong et al.}

\institute{MoE Key Lab of Artificial Intelligence, Shanghai Jiao Tong University, China \email{yanhong.sjtu@gmail.com, \{ustcnewly,c.sis\}@sjtu.edu.cn, zhang-lq@cs.sjtu.edu.cn}}
%\end{comment}
%******************

\maketitle

\begin{abstract}
Learning to generate new images for a novel category based on only a few images, named as few-shot image generation, has attracted increasing research interest. Several state-of-the-art works have yielded impressive results, but the diversity is still limited. In this work, we propose a novel Delta Generative Adversarial Network (DeltaGAN), which consists of a reconstruction subnetwork and a generation subnetwork. The reconstruction subnetwork captures intra-category transformation, \emph{i.e.}, ``delta'', between same-category pairs. The generation subnetwork generates sample-specific ``delta'' for an input image, which is combined with this input image to generate a new image within the same category. Besides, an adversarial delta matching loss is designed to link the above two subnetworks together. Extensive experiments on six benchmark datasets demonstrate the effectiveness of our proposed method. Our code is available at {https://github.com/bcmi/DeltaGAN-Few-Shot-Image-Generation}.
\end{abstract}
\section{Introduction}
 \label{sec:introduction}
\begin{figure}[t]
\begin{center}
\includegraphics[scale=0.07]{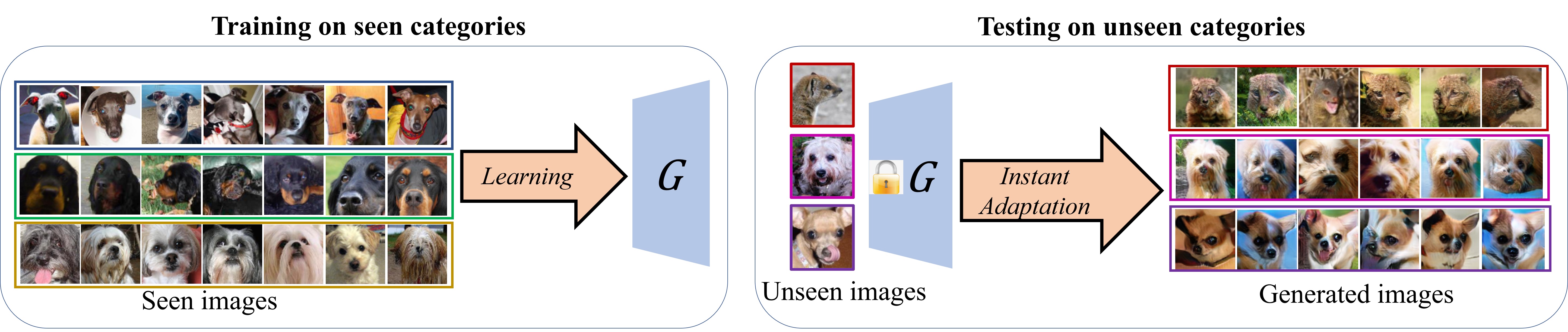}
\end{center}
\caption{{The illustration of few-shot image generation task. We train a generative model on multiple seen categories. The learned generative model can be instantly applied to generate new images for unseen categories at test time. Each color indicates one category}}
\label{fig:task_intro} 
\end{figure}

With the great success of deep learning, existing deep image generation models~\cite{stylegan1,stylegan2,binkowski2018demystifying,mao2017least,miyato2018spectral,brock2018large,donahue2019large,chen2019self,HoffmanTPZISED18,makhzani2017pixelgan} based on Variational Auto-Encoder (VAE)~\cite{vae} or Generative Adversarial Network (GAN)~\cite{goodfellow2014generative} have made a significant leap forward for generating diverse and realistic images for a given category.
{These methods generally require amounts of training images to generate new images for a given category. For the long-tail or newly emerging categories with only a few images, directly training or finetuning on limited data may cause overfitting issue~\cite{zhao2020leveraging,feng2019suppressing}.
Besides, it is very tedious to finetune the model for each unseen category.}
% However, these methods require amounts of training images to generate new images for a given category, which may fail in adapting to long-tail or newly emerging categories with only a few images.
% newly emerging unseen categories considering that labeling abundant images is time-consuming and expensive. 
Therefore, given a few images from an unseen category, it is necessary to consider how to generate new realistic and diverse images for this category instantly. This task is called few-shot image  generation in previous literature~\cite{bartunov2018few,antoniou2017data,hong2020matchinggan,hong2020f2gan}.
% {~\cite{ojha2021few,li2020few,robb2020few,WangGBHK020}, some of which ~\cite{ojha2021few,li2020few,robb2020few,WangGBHK020} focus on adapting the generative model pretrained on a large dataset to a small dataset with a few examples by finetuning process, while others~\cite{clouatre2019figr,hong2020matchinggan, hong2020f2gan} commit to achieve instant adaptation to unseen categories with a few examples without finetuning.
{In this paper, following ~\cite{bartunov2018few,antoniou2017data,hong2020matchinggan,hong2020f2gan}, we target at achieving instant adaptation from multiple seen categories to unseen categories without finetuning as shown in Fig.~\ref{fig:task_intro}, which can benefit a lot of downstream tasks like low-data classification and few-shot classification.}

The abovementioned few-shot image generation methods~\cite{bartunov2018few,antoniou2017data,hong2020matchinggan,hong2020f2gan} resort to seen categories with sufficient training images to train a generative model, which can be used to generate new images for an unseen category with only a few images, which are dubbed as conditional images. 
% In the training stage, one or more images from one seen category are fed into the model to generate new images for this seen category in each training episode.
% In the testing stage, given one or more images from each unseen category, the trained model can generate new images for this category. During both training and testing, the images fed into the model as sources to generate new images are dubbed as conditional images. 
For brevity, we refer to the images from seen (\emph{resp.}, unseen) categories as seen (\emph{resp.}, unseen) images. We classify the few-shot image generation methods into fusion-based methods~\cite{bartunov2018few,hong2020matchinggan,hong2020f2gan,gu2021lofgan} and transformation-based method~\cite{antoniou2017data}. However, those fusion-based methods can only produce images similar to conditional images and cannot be applied to one-shot image generation. Although transformation-based method could produce new images based on one conditional image, however, it fails to produce diverse images.

\begin{figure*}[t]
\begin{center}
\includegraphics[scale=0.11]{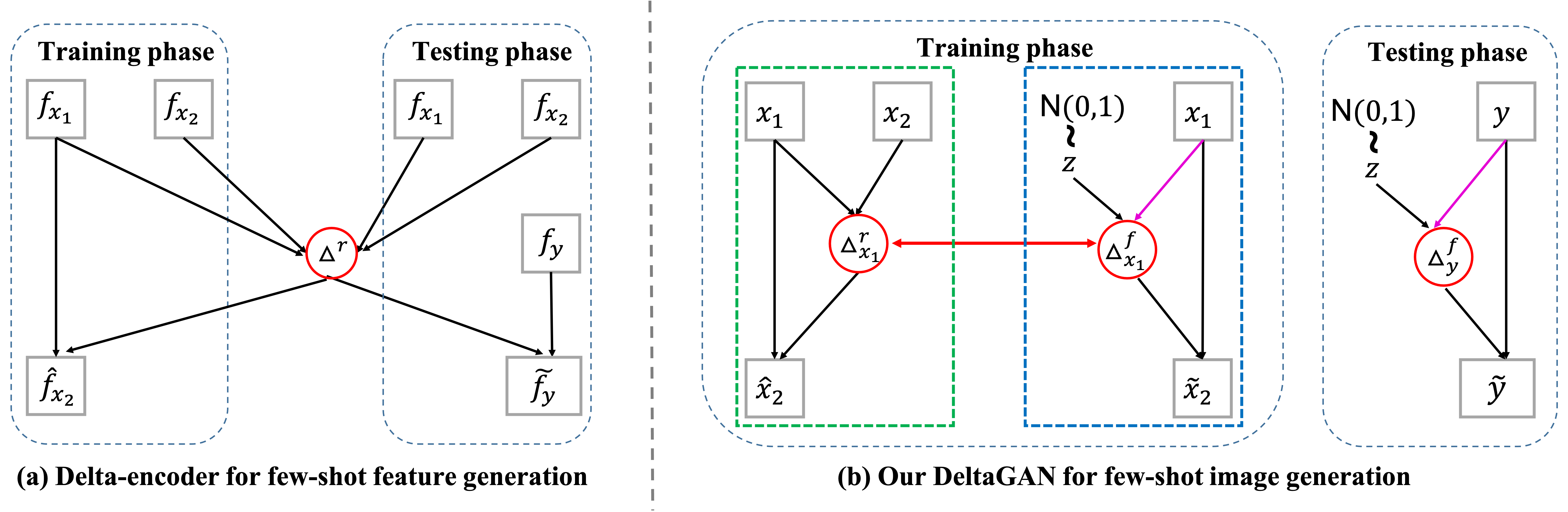}
\end{center}
\caption{The illustration of evolving from Delta-encoder to our DeltaGAN. $\{\bm{x}_1, \bm{x}_2\}$ (\emph{resp.,} $\{\bm{f}_{x_1}, \bm{f}_{x_2}\}$) is a same-category seen image pair (\emph{resp.,} feature pair). $\bm{y}$ (\emph{resp.,} $\bm{f}_y$ ) is a conditional image (\emph{resp.,} feature) from an unseen category.  $\{\hat{\bm{x}}_2$, $\tilde{\bm{x}}_2$, $\tilde{\bm{y}}\}$ (\emph{resp.,} $\{\hat{\bm{f}}_{x_2}$, $\tilde{\bm{f}}_y$\})  are generated images (\emph{resp.,} features). $\bm{z}$ is a random vector. 
%In (a), Delta-encoder extracts deltas from $\{X^S, Y^S\}$ and applies them to $X^U$ to generate new features belonging to the unseen category. In (b), the naive extension of Delta-encoder can extract deltas from seen category and map it into a random delta space to support stochastic sampling for new image generation. In (c), our DeltaGAN is proposed to learn sample-specific deltas to generate new images based on the design of the naive extension shown in (b).
$\{\bm{\Delta}^r$, $\bm{\Delta}_{x_1}^r\}$ (\emph{resp.}, $\{\bm{\Delta}_{x_1}^f$, $\bm{\Delta}_{y}^f\}$) means real (\emph{resp.}, fake) deltas. Red arrows indicate using adversarial delta matching loss to bridge the gap between real and fake delta. In (b), the green (\emph{resp.}, blue) box encloses the reconstruction (\emph{resp.}, generation) subnetwork, and pink arrows indicate the process of generating sample-specific delta
} 
\label{fig:comparison} 
\end{figure*}

Following the research line of transformation-based methods, we propose a novel Delta Generative Adversarial Network~(DeltaGAN), which can generate new images based on one conditional image by sampling random vectors. 
Our DeltaGAN is inspired by few-shot feature generation method Delta-encoder~\cite{schwartz2018delta}, in which intra-category transformation (\emph{i.e.}, the difference between two images within the same category) is called ``delta''. The main idea of Delta-encoder is shown in Fig.~\ref{fig:comparison}(a).
% Delta-encoder learns to apply transferable deformation feature to generate new features for a given unseen category. Specifically,
In the training stage, Delta-encoder learns to extract delta $\bm{\Delta}^r$ from same-category feature pair $\{\bm{f}_{x_1}, \bm{f}_{x_2}\}$ of image pair $\{\bm{x}_1, \bm{x}_2\}$ from seen categories, in which $\bm{\Delta}^r$ is the additional information required to reconstruct $\bm{f}_{x_2}$ from $\bm{f}_{x_1}$. We refer to $\bm{x}_1$ as conditional (source) sample and $\bm{x}_2$ as target sample.
In the testing stage, these extracted deltas are applied to a conditional feature $\bm{f}_y$ of image $\bm{y}$ from an unseen category to generate new feature $\tilde{\bm{f}}_y$ for this unseen category.
% Different from DAGAN~\cite{antoniou2017data} using random vectors as delta information, Delta-encoder can extract real delta information from same-category pair to generate new reasonable features. % 
However, Delta-encoder is a few-shot feature generation method, which cannot be directly applied to image generation. Besides, Delta-encoder relies on the deltas extracted from same-category training pairs, which does not support stochastic sampling (\emph{i.e.}, sampling random vectors) to generate new samples in the testing stage. 

In this paper, we aim to extend Delta-encoder to few-shot image generation method DeltaGAN, {which supports producing diverse deltas based on random vectors.
% which supports stochastic sampling in the testing stage. 
In this way, we can sample random vectors to generate diverse images without reaching training data in the testing stage. }
% One simple modification of Delta-encoder to support stochastic sampling is enforcing deltas to follow a prior distribution (\emph{e.g.}, unit Gaussian distribution) with KL divergence loss, so that we can sample random vectors from this prior distribution to generate new images in the testing stage. However, the delta between two images is very informative and imposing a strong prior on the delta may limit its information capacity~\cite{higgins2lerchner}, which leads to the failure of capturing informative deltas (see Section~\ref{sec:ablation}). 
% Therefore, we turn to learn a mapping from random vector to delta space as shown in Figure~\ref{fig:comparison}(b). One concern is that the plausibility of delta may depend on the conditional image~\cite{almahairi2018augmented}, that is, a plausible delta for one conditional image may be unsuitable for another conditional image (see Section~\ref{sec:ablation}). 
% Due to the above concerns,
Considering that the plausibility of delta may depend on the conditional image~\cite{almahairi2018augmented}, that is, a plausible delta for one conditional image may be unsuitable for another conditional image (see Section~\ref{sec:ablation}), we aim to produce sample-specific delta. 
In particular, we take in  a random vector and a conditional image to generate sample-specific delta, which represents the transformation from this conditional image to another possible image from the same category. We conjecture that the ability of generating sample-specific delta can be transferred from seen categories to unseen categories. 
To this end, we develop our DeltaGAN according to Fig.~\ref{fig:comparison}(b).
% and the intuition is also illustrated in Figure~\ref{fig:motivation_illustration}.
In the training phase, we use a reconstruction subnetwork to reconstruct $\bm{x}_2$ from $\bm{x}_1$ with the delta $\bm{\Delta}_{x_1}^r$  (real delta) extracted from $\{\bm{x}_1,\bm{x}_2\}$. We also use a generation subnetwork to generate sample-specific delta $\bm{\Delta}_{x_1}^f$ (fake delta) and produce new image $\tilde{\bm{x}}_2$.
To ensure that fake deltas function similarly to real ones, we introduce a novel adversarial delta matching loss by using a delta matching discriminator to judge whether an input-output image pair matches the corresponding delta.
Besides, we employ a variant of mode seeking loss~\cite{mao2019mode} to alleviate the mode collapse issue. 
We also employ typical adversarial loss and classification loss to make the generated images realistic and category-preserving.  In the testing stage, given a conditional unseen image $\bm{y}$, we can obtain its sample-specific delta $\bm{\Delta}_{y}^f$ by sampling random vector $\bm{z}$ for producing new image $\bm{\tilde{y}}$ from the category of $\bm{y}$.
% as shown in Figure~\ref{fig:motivation_illustration}(b)
Because each delta represents one possible intra-category transformation, given a conditional unseen image,  different deltas can produce realistic and diverse images from the same unseen category. Extensive experiments on six benchmark datasets demonstrate the effectiveness of our proposed method.
Our contributions can be summarized as follows: 
\begin{itemize}
\item We propose a novel delta-based few-shot image generation method, which has never been explored before.
\item Technically, we extend few-shot feature generation method Delta-encoder to few-shot image generation with stochastic sampling and sample-specific delta. We also design a novel adversarial delta matching loss.
\item Our method can produce diverse and realistic images for each unseen category based on a single conditional image, surpassing existing few-shot image generation methods by a large margin. 
\end{itemize}

% Our contributions can be summarized as follows: 1) We propose a novel delta-based few-shot image generation method, which has never been explored before.
% 2) Technically, we extend few-shot feature generation method Delta-encoder to few-shot image generation with stochastic sampling and sample-specific delta. We also design a novel adversarial delta matching loss.
% 3) Our method can produce diverse and realistic images for each unseen category based on a single conditional image, surpassing existing few-shot image generation methods by a large margin. 

% 1) we design an simple yet effective GAN for few-shot image generation; 2) Technically, In the generator, we construct a reconstruction module and a sampling module to map real sample-specific delta space to random sample-specific delta space to support stochastic sampling to generate more diverse images. In the discriminator, we propose a variant of mode seeking loss and a adversarial matching loss; 3) experiments on six real datasets demonstrate the effectiveness of our proposed method. 

\section{Related Work}
\label{sec:related}

\noindent\textbf{Data augmentation:}
% Data augmentation~\cite{Krizhevsky2012ImageNet} 
Data augmentation targets at augmenting training data with new samples. Traditional data augmentation tricks (\emph{e.g.}, crop, flip, color jittering) only have limited diversity. Also, there are some methods ~\cite{cubuk2019autoaugment,lim2019fast,ho2019population,tian2020improving}  proposed to learn optimal augmentation strategies to improve the accuracy of classifiers. Similarly, neural augmentation~\cite{perez2017the,ratner2017learning,jo2018bin,zhang2019dada,8509174} allowed a network to learn augmentations. As another research line, deep generative models can generate more diverse samples to augment training data, which can be categorized into feature-based augmentation methods~\cite{antoniou2017data} and image-based augmentation methods~\cite{schwartz2018delta}. Feature-based augmentation methods~\cite{chen2019multi,HariharanG17} focused on generating more diverse deep features to augment the feature space of training data, while image-based augmentation methods~\cite{chen2019image,tsutsui2019meta,hong2020matchinggan,hong2020f2gan} targeted at exploiting the distribution of training images and generating more diverse images.

\noindent\textbf{Few-shot feature generation}
In existing few-shot feature generation methods, the semantic knowledge learned from the seen categories is transferred to compensate unseen categories in~\cite{DixitKNV17,HariharanG17}. cCov-GAN~\cite{GaoSZZC18} proposed a covariance-preserving adversarial augmentation network to generate more features for unseen categories.
% The modes of variation from seen categories were transferred to unseen categories to hallucinate additional features for the unseen categories in \cite{HariharanG17}. 
In~\cite{WangGHH18}, a generator subnetwork was added to a classification network to generate new examples.
Intra-category diversity learned from seen categories was transferred to unseen categories to generate new features in~\cite{schwartz2018delta,9157338}.  Dual TriNet~\cite{chen2019multi} proposed to synthesize instance features by leveraging semantics using a novel auto-encoder network for unseen categories. {DTN~\cite{chen2020diversity} learned to transfer latent diversities from seen categories
and composite them with support features to generate diverse
features for unseen categories.
% which is optimized by minimizing an effective meta-classification loss in a single-stage network.
}

\noindent\textbf{Few-shot image generation}
Compared with few-shot feature generation, few-shot image generation is a more challenging problem.  Early methods can only be applied to generate new images for simple concepts, such as Bayesian program learning in~\cite{lake2011one}, Bayesian reasoning in~\cite{rezende2016one-shot}, and neural attention in~\cite{reed2018few-shot}.

Recently, several more advanced methods have been proposed to generate new real-world images in few-shot setting.
% Bayesian program learning was applied in ~\cite{lake2011one} to generate new images by learning simple concepts like pen stroke and combining the concepts hierarchically. Bayesian reasoning is used in ~\cite{rezende2016one-shot} to generate new images with a given simple concept by combining with representational power of deep learning. Neural attention was adopt in Attention PixelCNN~\cite{reed2018few-shot} for few-shot auto-regressive density modeling.
To name a few,  fusion-based method GMN~\cite{bartunov2018few} (\emph{resp.}, MatchingGAN~\cite{hong2020matchinggan}) combined  Matching Network~\cite{vinyals2016matching} with Variational Auto-Encoder~\cite{Pu2016Variational} (\emph{resp.}, Generative Adversarial Network~\cite{goodfellow2014generative}) to generate new images without finetuning in the test phase. F2GAN~\cite{hong2020f2gan} was designed to enhance the fusion ability of model by filling the details borrowed from conditional images. Transformation-based method DAGAN~\cite{antoniou2017data} proposed to produce new images by injecting random vectors into the generator conditioned on a single image. Apart from fusion-based and transformation-based methods, there also exist optimization-based methods. For example, FIGR~\cite{clouatre2019figr}  (\emph{resp.}, DAWSON~\cite{liang2020dawson}) combined adversarial learning with meta-learning method Reptile~\cite{nichol2018first}  (\emph{resp.}, MAML~\cite{finn2017model}) to generate new images. However, they need to fine-tune the trained model with unseen category. Moreover, they can hardly produce sharp and realistic images.
In this work, we propose a new transformation-based few-shot image generation method, which can produce more diverse images than previous methods based on a single image. 

% FIGR~\cite{clouatre2019figr} and DAWSON~\cite{liang2020dawson} adopted meta-learning algorithms to generate new images by fine-tuning the trained model with each unseen category, but they can hardly produce sharp and realistic images. 
{Note that some more recent works~\cite{ojha2021few,li2020few,robb2020few,WangGBHK020} are also called few-shot image generation. However, these works focus on adapting the generative model pretrained on a large dataset to a small dataset with a few examples, whose setting is quite different from ours. Firstly, these methods target at adapting from one source domain to another target domain, whereas our method adapts from multiple seen categories to unseen categories.  Secondly, the models of these works need to be finetuned for each unseen domain, which is very tedious. Instead, the model of our method can be instantly applied to unseen categories without finetuning. }

\begin{figure*}[t]
\begin{center}
\includegraphics[scale=0.19]{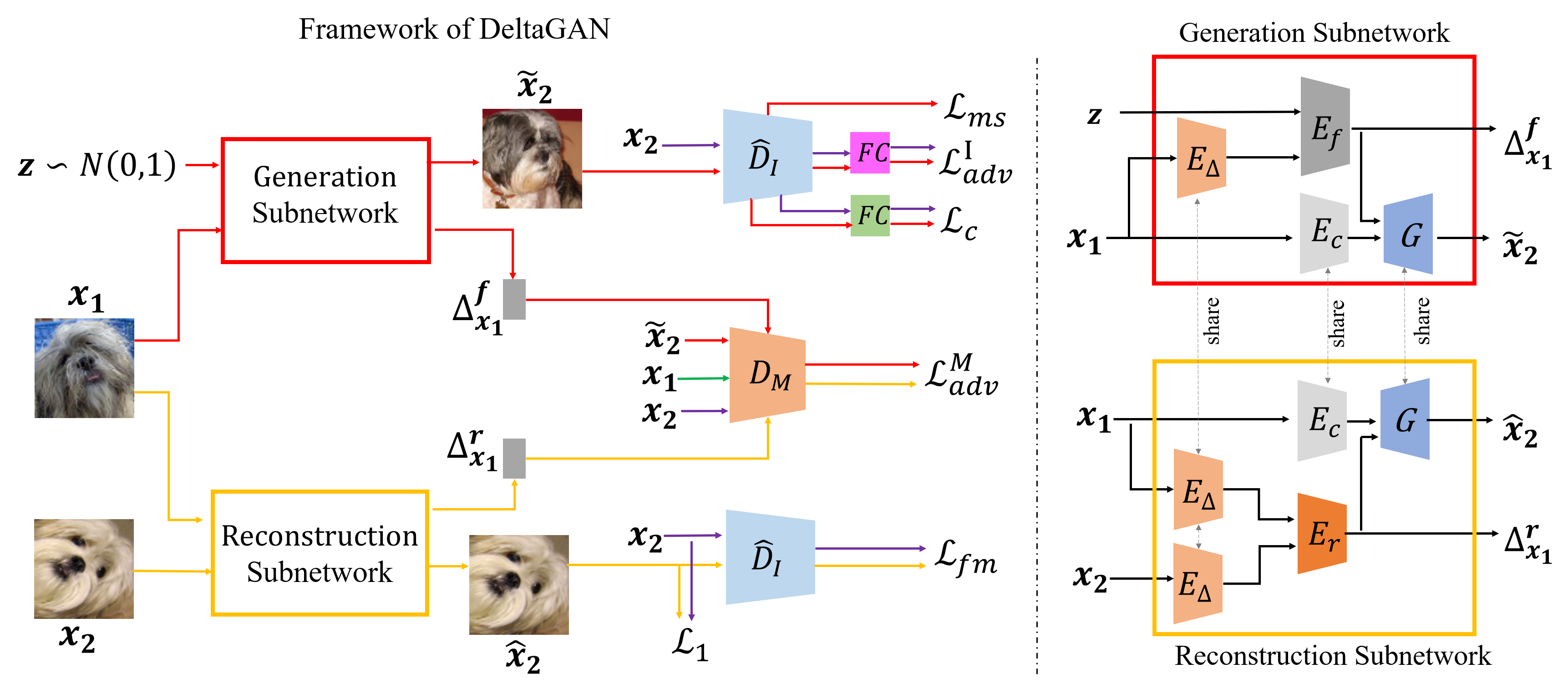}
\end{center}
\caption{Our DeltaGAN mainly consists of a reconstruction subnetwork and a generation subnetwork. Generation subnetwork learns to generate new image $\tilde{\bm{x}}_2$ based on conditional image $\bm{x}_1$ and random vector $\bm{z}$. Reconstruction subnetwork learns to produce reconstructed target image $\hat{\bm{x}}_2$ based on image pair $\{\bm{x}_1, \bm{x}_2\}$.  Best viewed in color
% At test time, given a unseen image $\bm{y}$, we only use generation subnetwork to produce diverse and realistic new images $\{ \tilde{\bm{y}}_k\}$ belonging to the category of $\bm{y}$. 
}
\label{fig:framework} 
\end{figure*}

\section{Our Method}  \label{sec:method}
We split all categories into seen categories and unseen categories, which have no overlap.  
Our DeltaGAN mainly consists of a reconstruction subnetwork and a generation subnetwork 
% and a delta matching discriminator 
as shown in Fig.~\ref{fig:framework}. The detailed architecture of each encoder/decoder is reported in Supplementary. In the training stage, given a same-category seen image pair $\{\bm{x}_1, \bm{x}_2\}$ where $\bm{x}_1$ is the conditional image and $\bm{x}_2$ is the target image, the reconstruction subnetwork extracts real delta $\bm{\Delta^{r}_{x_1}}$ from this pair, and reconstructs the target image $\bm{x}_2$ based on $\bm{x}_1$ and $\bm{\Delta^{r}_{x_1}}$. In the generation subnetwork, a random vector $\bm{z}$ and the conditional image  $\bm{x}_1$ are used to obtain fake sample-specific delta  $\bm{\Delta^{f}_{x_1}}$, which collaborates with $\bm{x}_1$ to generate a new image $\tilde{\bm{x}}_2$. Moreover, we design an adversarial delta matching loss to  bridge the gap between real delta and fake delta. {In the testing stage, given an unseen image $\bm{y}$,  only generation subnetwork is used to produce diverse and realistic images $\{\tilde{\bm{y}}_k\}$ belonging to the same category of $\bm{y}$.}

% \begin{figure}
% \begin{center}
% \includegraphics[scale=0.42]{./figures/attention.png}
% \end{center}
% \caption{The architecture of our Non-local ATtention (NAT) module. $\bm{{\xi}}_r^k$ is the feature of $\bm{x}_k$ from the $r$-th encoder block, $\bm{{\phi}}_r$ is the output from the $r$-th decoder block, and $\bm{{\eta}}_r$ is the output of NAT. }
% \label{fig:attention} 
% \end{figure}

% \subsection{Delta Generator}  \label{sec:generator}
% Our delta generator $G$ is constructed by a reconstruction subnetwork $R$ and a stochastic generation subnetwork $S$ shown in Figure~\ref{fig:framework}. 

\subsection{Reconstruction Subnetwork}\label{sec:rec_network}
% \noindent\textbf{Reconstruction subnetwork:} 

In the reconstruction subnetwork (see Fig.~\ref{fig:framework}), there are three encoders $E_{\Delta}$, $E_c$, $E_r$ and a decoder $G$. Given a same-category seen image pair $\{\bm{x}_1, \bm{x}_2\}$,
% our reconstruction subnetwork $R$ is designed to 
we use $E_{\Delta}$ to extract paired features $\{E_{\Delta} (\bm{x}_1), E_{\Delta} (\bm{x}_2)\} \in \mathcal{R}^{W \times H \times C}$, where $W \times H$ denotes the feature map size and $C$ denotes the channel number. Then, we calculate the difference between $E_{\Delta} (\bm{x}_2)$ and $E_{\Delta} (\bm{x}_1)$, which is fed into $E_r$ to obtain real %sample-specific 
delta $\bm{\Delta}_{x_1}^{r} \in \mathcal{R}^{W \times H \times C}$:
\begin{equation}
\begin{aligned}
\bm{\Delta}_{x_1}^{r} = E_r(E_{\Delta} (\bm{x}_2) - E_{\Delta} (\bm{x}_1)), \label{real_delta}
\end{aligned}
\end{equation}
where $\bm{\Delta}_{x_1}^{r}$ contains the additional information needed to reconstruct $\bm{x}_2$ from $\bm{x}_1$. We do not restrict our delta features to be linear offsets, which enables the delta features to learn more complex transformations. Then,  $\bm{\Delta}_{x_1}^{r}$ is concatenated with $E_c (\bm{x}_1) \in \mathcal{R}^{W \times H \times C}$ and fed into $G$ to obtain the reconstructed image $\hat{\bm{x}}_2$:
\begin{equation}
\begin{aligned}
\hat{\bm{x}}_2 = G (\bm{\Delta}_{x_1}^{r}, E_{c}(\bm{x}_1)). \label{reronc}
\end{aligned}
\end{equation}
%$\hat{\bm{x}}_2$ is the reconstructed result by applying extracted $\bm{\Delta}_{x_1}^{r}$ to condition image $\bm{x}_1$. To ensure the reconstruction ability of our reconstruction subnetwork, 
We employ a reconstruction loss $\mathcal{L}_1$ to ensure that $\hat{\bm{x}}_2$ is close to $\bm{x_2}$:
\begin{equation} \label{eqn:loss_reconstruction}
\begin{aligned}
\mathcal{L}_1 =  || \hat{\bm{x}}_2 - \bm{x}_2 ||_1.
\end{aligned}
\end{equation}
Considering the instability issue of early training stage, we use a feature matching loss~\cite{BaoCWLH17} by matching the discriminative feature of $\hat{\bm{x}}_2$ with that of $\bm{x}_2$. In detail, we use a feature extractor $\hat{D}_{I}$ to extract the discriminative features of $\hat{\bm{x}}_2$ and $\bm{x}_2$ in each layer to calculate the feature matching loss:
\begin{equation} \label{eqn:feature_mmatching_loss}
\begin{aligned}
\mathcal{L}_{fm} =   \frac {1}{L}  \sum  _{l=1}^{L}  || \hat{D}_I^l(\bm{x}_2) - \hat{D}_I^l(\hat{\bm{x}}_2)||_1,
\end{aligned}
\end{equation}
where $L$ is the layer number of $\hat{D}_{I}$.

% \input{sections/figures_tables/framework_detail}
% \noindent\textbf{Generation subnetwork:}

% To map extracted sample-specific delta into a random sample-specific delta space, 
To support stochastic sampling for generation, we design another generation subnetwork in parallel with the reconstruction subnetwork (see Fig.~\ref{fig:framework}). Two subnetworks share two encoders $E_{\Delta}$, $E_c$ and the decoder $G$. Besides, a new encoder $E_f$ is introduced to obtain fake sample-specific delta. 
In our generation subnetwork, we concatenate a random vector $\bm{z}$ sampled from unit Gaussian distribution and the feature of conditional image $E_{\Delta}(\bm{x}_1) \in \mathcal{R}^{W \times H \times C} $, which is fed into $E_f$ to obtain sample-specific delta $\bm{\Delta}_{x_1}^{f} \in \mathcal{R}^{W \times H \times C}$:
\begin{equation}
\begin{aligned}
\bm{\Delta}_{x_1}^{f} = E_f(\bm{z}, E_{\Delta} (\bm{x}_1) ), \label{fake_delta}
\end{aligned}
\end{equation}
where $\bm{\Delta}_{x_1}^{f}$ contains the additional information needed to transform conditional image $\bm{x}_1$ to another possible image within the same category. Then, analogous to the reconstruction subnetwork, $\bm{\Delta}_{x_1}^{f}$ is concatenated with $E_c (\bm{x}_1)$ and fed into $G$ to produce a new image $\tilde{\bm{x}}_2$ belonging to the category of $\bm{x}_1$:
\begin{equation}
\begin{aligned}
\tilde{\bm{x}}_2 = G (\bm{\Delta}_{x_1}^{f}, E_c(\bm{x}_1)), \label{sampling}
\end{aligned}
\end{equation}
in which $\tilde{\bm{x}}_2$ is the transformed result after applying delta $\bm{\Delta}_{x_1}^{f}$ to $\bm{x}_1$. %To ensure the fidelity and diversity of generated images, we use a adversarial learning fashion by discriminator, which is detailed in the next section.

% \subsection{Adversarial Mode Seeking Discriminator}\label{sec:other_loss}
% \subsection{Mode Seeking Discriminator}\label{sec:other_loss}
%To generate diverse images belonging to the same category of given condition image, we design an auxiliary-assisted discriminator $\mathrm{D}_{I}$ to seek minor mode and classify generated images into correct category beyond distinguishing real image $\bm{x}_2$ from fake images $\tilde{\bm{x}}_2$. The architecture of $\mathrm{D}_{I}$  can be found in Supplementary.
\subsection{Generation Subnetwork}\label{sec:gen_network}

\noindent\textbf{Adversarial loss: }To make the generated image $\tilde{\bm{x}}_2$ close to real images, we employ a standard adversarial loss using the discriminator $\mathrm{D}_{I}$. $\mathrm{D}_{I}$ contains the feature extractor $\hat{D}_{I}$ mentioned in Section~\ref{sec:rec_network} and a fully-connected (FC) layer. 
%Firstly, we use $\mathrm{D}_{I}$ to calculate the discriminative scores $\mathrm{D}_{I}(\bm{x}_2)$  and $\mathrm{D}_{I}(\tilde{\bm{x}}_2)$, which are used for adversarial learning. Concretely, 
We adopt the hinge adversarial loss proposed in~\cite{miyato2018cgans}:
\begin{eqnarray} \label{eqn:adversarial_image_loss}
\mathcal{L}_{adv,D}^{I} && = \mathbb{E}_{\bm{x}_2}  [\max (0,1\!-\!\mathrm{D}_{I}({\bm{x}}_2))] \!+\! \mathbb{E}_{\tilde{\bm{x}}_2} [\max (0,1\!+\!\mathrm{D}_{I}(\tilde{\bm{x}}_2))], \nonumber\\ \mathcal{L}_{adv,G}^{I} &&= -\mathbb{E}_{\tilde{\bm{x}}_2} [\mathrm{D}_{I}(\tilde{\bm{x}}_2)].
\end{eqnarray}
The discriminator $\mathrm{D}_{I}$ tends to distinguish fake images from real images by minimizing $\mathcal{L}_{adv,D}^{I}$, while the generator tends to generate realistic images to fool the discriminator by minimizing $\mathcal{L}_{adv,G}^{I}$.

\noindent\textbf{Classification loss: }To ensure that $\tilde{\bm{x}}_2$ belongs to the expected category, we construct a classifier by replacing the last FC layer of $D_{I}$ with another FC layer (the number of outputs is the number of seen categories).
% , analogous to ACGAN~\cite{odena2017conditional}. 
Then, the images from different categories can be distinguished by a cross-entropy classification loss:
\begin{equation}
\begin{aligned}\label{eqn:loss_classification}
\mathcal{L}_{c} = -\log p(c(\bm{x})|\bm{x}),
\end{aligned}
\end{equation}
where $c(\bm{x})$ is the category label of $\bm{x}$. We train the classifier by minimizing $\mathcal{L}_{c,D} = -\log p(c(\bm{x}_2)|\bm{x}_2)$ of the target image $\bm{x}_2$.  We also expect the generated image $\tilde{\bm{x}}_2$ to be classified as the same category of target image $\bm{x}_2$. Thus, we minimize $\mathcal{L}_{c,G}=-\log p(c(\bm{x}_2)|\tilde{\bm{x}}_2)$ when updating the generator.

\noindent\textbf{Adversarial delta matching loss: }
To ensure that the generated sample-specific deltas function similarly to real deltas and encode the intra-category transformation, we design a novel adversarial delta matching loss to bridge the gap between real deltas and fake deltas. This goal is accomplished by a delta matching discriminator $\mathrm{D}_{M}$, which takes a triplet (conditional image, output image, the delta between them) as input as shown in Fig.~\ref{fig:framework}. Our delta matching discriminator $\mathrm{D}_{M}$ is constructed by feature extractor $\mathrm{\hat{D}}_{I}$ and four FC layers following global average pooling. In delta matching discriminator $\mathrm{D}_{M}$, we extract the features of paired images $\{\mathrm{\hat{D}}_{I}(\bm{x}_1), \mathrm{\hat{D}}_{I}(\bm{x}_2)\}$ (\emph{resp.}, $\{\mathrm{\hat{D}}_{I}(\bm{x}_1), \mathrm{\hat{D}}_{I}(\tilde{\bm{x}}_2)\}$), which are concatenated with sample-specific delta $\bm{\Delta}_{x_1}^{r}$(\emph{resp.}, $\bm{\Delta}_{x_1}^{f}$) to form a real (\emph{resp.}, fake) triplet.
Then, the real triplet and fake triplet are fed into the four FC layers to judge whether this conditional-output image pair matches the corresponding delta, in other words, whether the delta is the additional information required to transform the conditional image to the output image. 
In adversarial learning, the discriminator $\mathrm{D}_{M}$ needs to distinguish the real triplet $\{\bm{x}_1,  \bm{x}_2, \bm{\Delta}^r_{x_1}\}$ from the fake triplet $\{\bm{x}_1, \tilde{\bm{x}}_2, \bm{\Delta}^f_{x_1}\}$, while the generator aims to synthesize realistic fake triplet to fool the discriminator. The delta matching adversarial loss is also in the form of hinge adversarial loss~\cite{miyato2018cgans}, which can be written as
\begin{eqnarray} \label{eqn:adversarial_delta_loss}
\mathcal{L}_{adv,D}^{M}  &&=\! \mathbb{E}_{ \bm{x}_1, {\bm{x}}_2, {\bm{\Delta}}^r_{x_1} }  [\max(0, 1 \!-\! \mathrm{D}_{M}(\bm{x}_1, \bm{x}_2, {\bm{\Delta}}^r_{x_1}))] \nonumber\\
&&+\mathbb{E}_{ \bm{x}_1, \tilde{\bm{x}}_2, {\bm{\Delta}}^f_{x_1} } [\max (0,1 + \mathrm{D}_{M}(\bm{x}_1, \tilde{\bm{x}}_2, {\bm{\Delta}}^f_{x_1}))],  \nonumber\\
\mathcal{L}_{adv,G}^{M} &&= -\mathbb{E}_{ \bm{x}_1, \tilde{\bm{x}}_2, {\bm{\Delta}}^f_{x_1} }[\mathrm{D}_{M}(\bm{x}_1, \tilde{\bm{x}}_2, \bm{\Delta}^f_{x_1})],
\end{eqnarray}
where $\mathcal{L}_{adv,D}^{M}$ (\emph{resp.}, $\mathcal{L}_{adv,G}^{M}$) is optimized for updating  $\{\hat{\mathrm{D}}_{I},\mathrm{D}_{M}\}$ (\emph{resp.}, the generator).

\noindent\textbf{Mode seeking loss: }
We observe that by varying random vector $\bm{z}$, the generated images may collapse into a few modes, which is referred to as mode collapse~\cite{mao2019mode}. 
Therefore, we use a variant of mode seeking loss~\cite{mao2019mode} to seek for more modes to enhance the diversity of generated images. Different from~\cite{mao2019mode}, we apply mode seeking loss to multi-layer features extracted by $\hat{D}_{I}$. In particular, we minimize the ratio of the distance between $\bm{z}_1$ and $\bm{z}_2$ over the distance between $\hat{D}_{I}^l(\tilde{\bm{x}}_2^1)$ and $\hat{D}_{I}^l(\tilde{\bm{x}}_2^2)$ at the $l$-th layer of $\hat{D}_{I}$:
\begin{eqnarray}  \label{eqn:mode_seeking_loss}
\!\!\!\!\!\!\!\!&&\mathcal{L}_{ms} =  \frac {1}{L}   \sum  _{l=1}^{L}   \frac {|| \bm{z}_1 - \bm{z}_2||_1} {||\hat{D}^l_{I}(\tilde{\bm{x}}_2^1) -  \hat{D}^l_{I}(\tilde{\bm{x}}_2^2)||_1}.
\end{eqnarray}
Intuitively, when $|| \bm{z}_1 - \bm{z}_2||_1$ is large, we expect $\hat{D}^l_{I}(\tilde{\bm{x}}_2^1)$ and $\hat{D}^l_{I}(\tilde{\bm{x}}_2^2)$ to be considerably different, which can push the generator to search more modes to produce diverse images. In our experiments (see Section~\ref{sec:ablation}), we find that mode seeking loss is critical for diversity. However, without the guidance of reconstruction subnetwork and adversarial delta matching loss, solely using mode seeking loss cannot generate meaningful deltas, with both diversity and realism significantly downgraded.

% but only on the premise that the generated delta $\bm{\Delta}_{x_1}^{f}$ is close to real delta. 

% Without the guidance of reconstruction subnetwork and adversarial delta matching loss, solely using mode seeking loss cannot bring much diversity (see Section~\ref{sec:ablation}). 

% \noindent\textbf{Mode seeking discriminator:} 
% \subsection{Other Losses}\label{sec:other_loss}

\subsection{Optimization}
We use $\theta_G$ to denote the model parameters of $\{E_{\Delta}, E_r, E_c, E_f, G\}$, while $\theta_D$ is used to denote the model parameters of $\{D_I,D_M\}$.
The total loss function of our method can be written as 
\begin{equation} \label{optimization}
\begin{aligned}
\mathcal{L} =  \mathcal{L}_{adv}^I +  \mathcal{L}_{adv}^M  + \lambda_1 \mathcal{L}_{1}  + \mathcal{L}_{c} + \lambda_{fm} \mathcal{L}_{fm} + \lambda_{ms} \mathcal{L}_{ms},
\end{aligned}
\end{equation}
in which $\lambda_1$, $\lambda_{fm}$, and $\lambda_{ms}$ are trade-off parameters. $\mathcal{L}_{adv}^I$ represents $\mathcal{L}_{adv,G}^I$ (\emph{resp.}, $\mathcal{L}_{adv,D}^I$) when updating the model parameters $\theta_G$ (\emph{resp.}, $\theta_D$). Similarly, $\mathcal{L}_{adv}^M$ represents $\mathcal{L}_{adv,G}^M$ (\emph{resp.}, $\mathcal{L}_{adv,D}^M$) when updating the model parameters $\theta_G$ (\emph{resp.}, $\theta_D$).
%In the process of adversarial learning, delta generator and delta discriminator are optimized by related loss terms in an alternating manner. 
% $\theta_{E_\Delta}$ (\emph{resp.},  $\theta_{E_c}$,  $\theta_{E_r}$,  $\theta_{E_s}$,  $\theta_{G_c}$,  $\theta_{D}$ ) to denote the model parameters of each encoder and decoder. Use \theta_G, \theta_D to group the model parameters (e.g., \theta_G={XXX, XXXX}). In each step, clarify which model parameters should be optimized. 

$\theta_G$ and $\theta_D$ are optimized using related loss terms in an alternating fashion. 
In particular, $\theta_D$ is optimized by minimizing $\mathcal{L}_{adv,D}^I+\mathcal{L}_{adv,D}^M+\mathcal{L}_{c,D}$.
% similar to ACGAN~\cite{odena2017conditional}. 
$\theta_G$ is optimized by minimizing $\mathcal{L}_{adv,G}^I+\mathcal{L}_{adv,G}^M+ \lambda_1\mathcal{L}_{1}+ \mathcal{L}_{c,G}+\lambda_{fm}\mathcal{L}_{fm}+\lambda_{ms}\mathcal{L}_{ms}$, in which $\mathcal{L}_{c,D}$ and $\mathcal{L}_{c,G}$ are defined below Eqn.~\ref{eqn:loss_classification}.

\section{Experiments}\label{sec:experiments}
We conduct experiments on six few-shot image datasets: 
EMNIST~\cite{cohen2017emnist},  
VGGFace~\cite{cao2018vggface2}, Flowers~\cite{nilsback2008automated}, Animal Faces~\cite{deng2009imagenet}, NABirds~\cite{van2015building}, and Foods~\cite{kawano2014automatic}. Following MatchingGAN and FUNIT, we split all categories into seen categories and unseen categories. After having a few trials, we set $\lambda_1=10$, $\lambda_{fm} = 0.1$, and $\lambda_{ms} = 10$ by observing the quality of generated images during training. We adopt Adam optimizer with learning rate of $1e-4$. The batch size is set to $16$ and our model is trained for $200$ epochs. 
The details of datasets and implementation are reported in Supplementary.

\setlength{\tabcolsep}{1.2pt}
\begin{table}[t]
  \caption{FID ($\downarrow$) and LPIPS ($\uparrow$) of images generated by different methods for unseen categories on four datasets in 1/3-shot setting 
%   (from left to right:VGGFace, Flowers, Animal Faces, and NABirds).
} 
  \centering
  %\fontsize{8}{8}\selectfont
  \resizebox{0.9\columnwidth}{!} {
  \begin{tabular}{l|c|cc|cc|cc|cc}
      % \Xhline{1.2pt}
      \toprule[0.8pt]
      \multirow{2}{*}{Method} &
      \multirow{2}{*}{Shot}  &
      \multicolumn{2}{c|}{VGGFace} &
      \multicolumn{2}{c|}{Flowers} &
      \multicolumn{2}{c|}{Animal Faces} &
      \multicolumn{2}{c}{NABirds}  \cr
      & &FID   &  LPIPS  & FID   & LPIPS  &FID  & LPIPS & FID   & LPIPS \cr 
      \cmidrule(r){1-1}   \cmidrule(r){2-2}  \cmidrule(r){3-4}  \cmidrule(r){5-6}  \cmidrule(r){7-8} \cmidrule(r){9-10}
    FIGR \cite{clouatre2019figr} &3&139.83 &0.0834 & 190.12 &0.0634 &211.54  &0.0756 & 210.75&0.0918 \cr
    DAWSON \cite{liang2020dawson} &3&137.82  & 0.0769 &  188.96 &0.0583  & 208.68  &0.0642 &181.97 &0.1105 \cr
    GMN \cite{bartunov2018few} &3&136.21  &0.0902 &200.11 &0.0743 &220.45  &0.0868 & 208.74&0.0923  \cr
    % DAGAN~\cite{antoniou2017data}& 121.43 & 4.12 &0.08 & 142.65& 2.81& 0.08&152.11 &2.74 &0.09\cr
    DAGAN \cite{antoniou2017data} &3& 128.34  & 0.0913& 151.21&0.0812 &155.29 &0.0892 & 159.69 &0.1405 \cr
    DAGAN \cite{antoniou2017data} &1& \textit{134.28} &\textit{0.0608}  &\textit{179.59} &\textit{0.0496} & \textit{185.54} &\textit{0.0687} & \textit{183.57} &\textit{0.0967}\cr
    MatchingGAN\cite{hong2020matchinggan} &3& 118.62  & 0.1695&  143.35&0.1627 & 148.52& 0.1514& 142.52 &0.1915  \cr
    F2GAN \cite{hong2020f2gan} &3&109.16 & 0.2125 &120.48 &0.2172 &117.74  &0.1831 &126.15 &0.2015 \cr
    LoFGAN \cite{gu2021lofgan}&3&106.24 & 0.2096 &112.55 &0.2687 &116.45 &0.1756 &124.56 &0.2041  \cr
    \midrule
    
    DeltaGAN &3&$\textbf{78.35}$ & $\textbf{0.3487}$ &$\textbf{104.62}$ & $\textbf{0.4281}$& $\textbf{87.0}$4 & $\textbf{0.4642}$& $\textbf{95.97}$&$\textbf{0.5136}$ \cr
    DeltaGAN &1& $\textbf{\textit{80.12}}$&$\textbf{\textit{0.3146}}$  &$\textbf{\textit{109.78}}$ &$\textbf{\textit{0.3912}}$ &$\textbf{\textit{89.81}}$  & $\textbf{\textit{0.4418}}$&$\textbf{\textit{96.79}}$ &$\textbf{\textit{0.5069}}$ \cr
    % \Xhline{1.2pt}
    \bottomrule[0.8pt]
    
  \end{tabular}
  }
%   \vspace{0.1mm}
  \label{tab:performance_metric}
\end{table}
\begin{figure}[t]
\begin{center}
\includegraphics[scale=0.18]{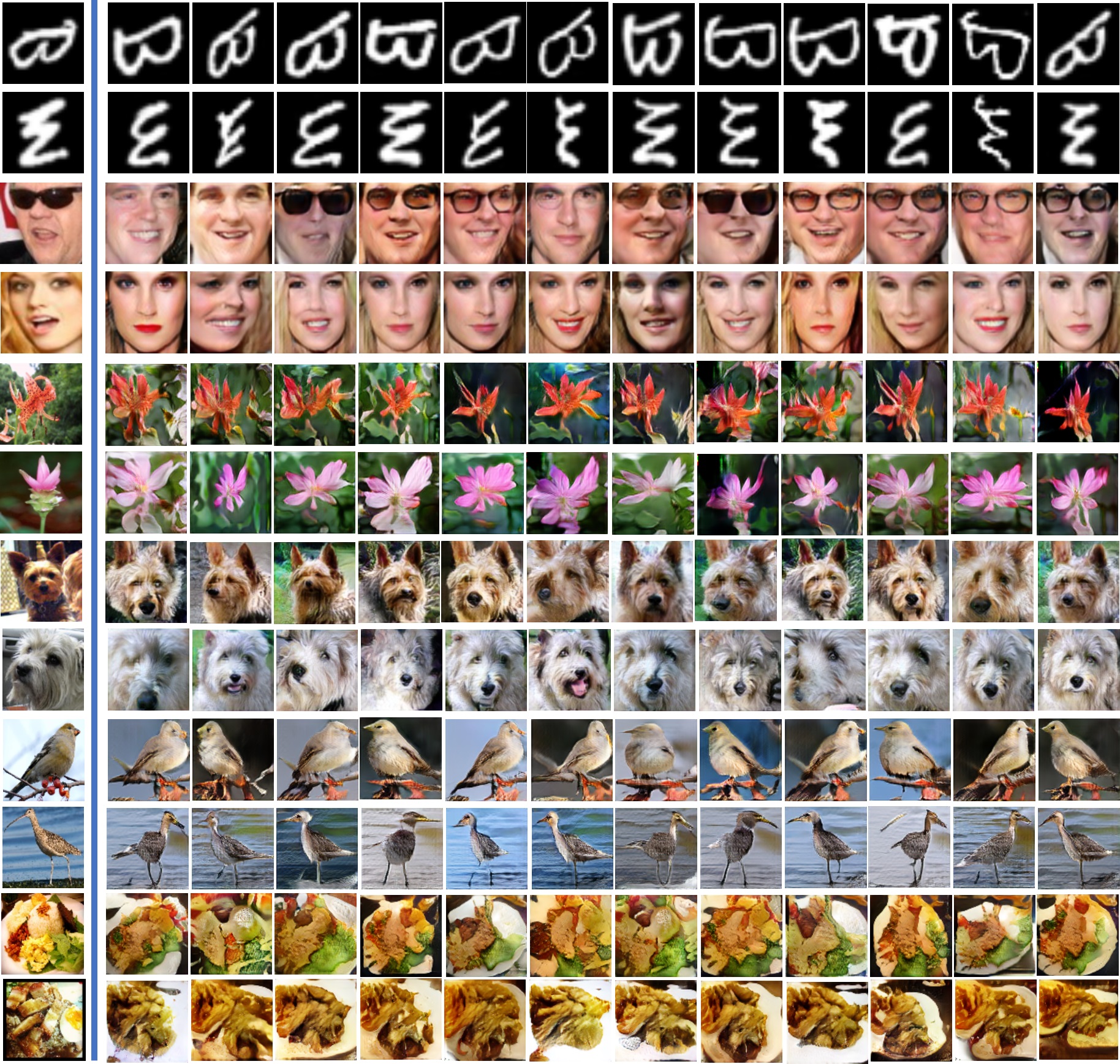}
\end{center}
\caption{Images generated by our DeltaGAN in 1-shot setting on four datasets (from top to bottom:EMNIST, VGGFace, Flowers, Animal Faces, NABirds, and Foods). The conditional images are in the leftmost column}
\label{fig:visualization} 
\end{figure}

\begin{figure*}[t]
\begin{center}
\includegraphics[scale=0.14]{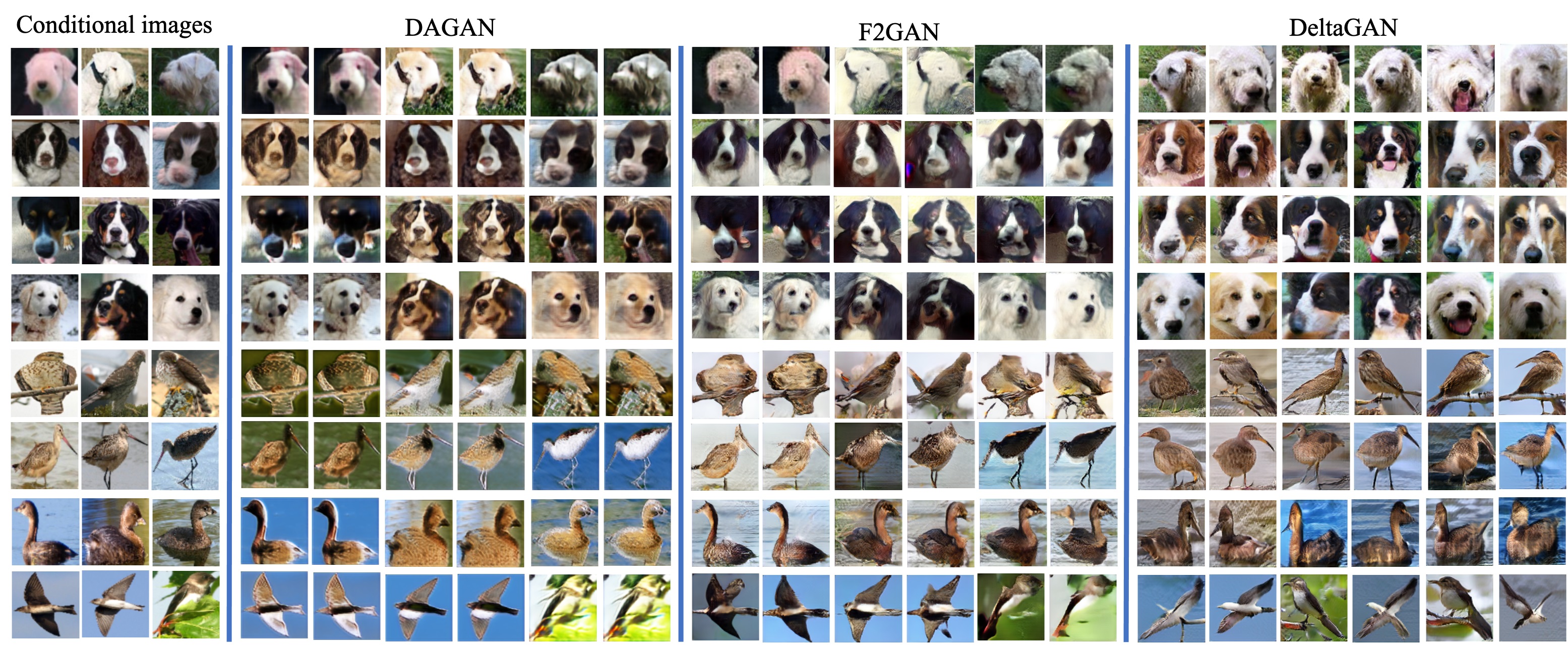}
\end{center}
\caption{Images generated by DAGAN, F2GAN, and our DeltaGAN in $3$-shot setting on two datasets (from top to bottom: Animal Faces and NABirds). The conditional images are in the left three columns}
\label{fig:visualization_compare} 
\end{figure*}

\subsection{Evaluation of Generated Images} \label{sec:visualization}
To evaluate the quality of images generated by different methods, 
we calculate Fréchet Inception Distance (FID)~\cite{heusel2017gans} and Learned Perceptual Image Patch Similarity (LPIPS)~\cite{zhang2018unreasonable} on four datasets. 
FID is used to measure the distance between the extracted features of generated unseen images and those of real unseen images. LPIPS is used to measure the diversity of generated unseen images. For each unseen category, the average of pairwise distances among generated images is calculated, and then the average of all unseen categories is calculated as the final LPIPS score. Since the number of conditional images in fusion-based methods GMN~\cite{bartunov2018few}, MatchingGAN~\cite{hong2020matchinggan}, F2GAN~\cite{hong2020f2gan}, and LoFGAN~\cite{gu2021lofgan}) is a tunable hyper-parameter, we use 3 conditional images in each training and testing episode. %, and we can use those trained models to generate images for unseen categories in testing phase conditioned on $3$ conditional images. 
In the testing stage, if $K$ images are provided for each unseen category, we refer to this setting as $K$-shot setting. We report the $3$-shot results for all methods and $1$-shot results for the methods which only require one conditional image. 

In either setting, following \cite{gu2021lofgan,hong2020f2gan}, we use each method to generate $128$ images for each unseen category, which are used to calculate FID and LPIPS. For DeltaGAN and DAGAN which are applicable to both $1$-shot and $3$-shot settings, we generate $128$ images based on one conditional image in $1$-shot setting and generate $128$ images by randomly sampling one conditional image each time in $3$-shot setting. The results are summarized in Table \ref{tab:performance_metric}, we can observe that our method achieves the lowest FID and highest LPIPS in the $3$-shot setting, which demonstrates that our method could generate more diverse and realistic images compared with baseline methods. Besides, our method in $1$-shot setting also achieves competitive results, which are even better than other baselines in $3$-shot setting. We also compare our DeltaGAN with other few-shot image generation method~\cite{ojha2021few} in Supplementary.

We show some example images generated by our DeltaGAN on six datasets in Fig.~\ref{fig:visualization}. We exhibit $12$ generated images based on one conditional unseen image by sampling different random vectors. 
On EMNIST dataset, we can see that generated images maintain the concepts of conditional images and have remarkable diversity. 
On natural datasets VGGFace, Flowers, Animal Faces, NABirds, and Foods, our DeltaGAN can generate diverse images with high fidelity. 

For comparison, we also show some example images generated by DAGAN and F2GAN in Fig.~\ref{fig:visualization_compare}. For DAGAN, we arrange the results according to the conditional image. It can be seen that the structures of images produced by DAGAN are almost the same as the conditional image. For F2GAN, the generated images are still close to one of the conditional images and may have unreasonable shapes when fusing conditional images.
Apparently, our DeltaGAN can produce images of higher quality and more diversity.

\subsection{Few-shot Classification}
\label{sec:few-shot}
In this section, we demonstrate that the new images generated by our DeltaGAN can greatly benefit few-shot classification. 
The experiments for low-data classification and comparison with traditional data augmentation methods can be found in Supplementary.
Following the $N$-way $C$-shot setting in few-shot classification~\cite{finn2017model}, in which evaluation episodes are created and the averaged accuracy over multiple evaluation episodes is calculated for evaluation. In each evaluation episode, $N$ categories from unseen categories are randomly selected and $C$ images from each of $N$ categories are randomly selected. These selected $N\times C$ images are used as training set while the remaining unseen images from $N$ unseen categories are used as test set. We pretrain ResNet$18$~\cite{he2016deep} on the seen images and remove the last FC layer as the feature extractor, which is used to extract features for unseen images. 
In each evaluation episode in $N$-way $C$-shot setting, our DeltaGAN generates $512$ new images to augment each of $N$ categories. Based on the extracted features, we train a linear classifier with $N\times(C+512)$ training images, which is then applied to the test set. 
we train a linear classifier to evaluate the few-shot generation ability of our DeltaGAN. Besides $N\times C$ training images, our generator can generate $512$ images to augment each of $N$ categories in the training set. 
\setlength{\tabcolsep}{3pt}
\begin{table}[t]
  \caption{Accuracy($\%$) of different methods on three datasets in few-shot classification setting ($10$-way $1/5$-shot). Note that fusion-based methods MatchingGAN, F2GAN, and LoFGAN are not applicable in $1$-shot setting} 
  \centering
  \resizebox{0.7\columnwidth}{!} {
  %\fontsize{8}{8}\selectfont
  \begin{tabular}{l|rr|rr|rr}
      % \Xhline{1.2pt} 
      \toprule[0.8pt]
      \multirow{2}{*}{Method}&\multicolumn{2}{c|}{VGGFace}&\multicolumn{2}{c|}{Flowers}&\multicolumn{2}{c}{Animal Faces}
      \cr & 1-shot &5-shot &1-shot & 5-shot & 1-shot & 5-shot\cr
      \cmidrule(r){1-1}  \cmidrule(r){2-3}  \cmidrule(r){4-5}  \cmidrule(r){6-7}
    MatchingNets \cite{vinyals2016matching} & 33.68 &48.67  &40.96 &56.12  & 36.54 &50.12 \cr

    MAML \cite{finn2017model} &32.16 &47.89   &42.95 &58.01  &35.98   &49.89 \cr

    RelationNets \cite{sung2018learning} &39.95 & 54.12 &48.18 &61.03  & 45.32 & 58.12 \cr

    MTL \cite{sun2019meta}& 51.45 &68.95  &54.34 &73.24  &52.54  &70.91 \cr

    % DN4&52.88 &70.02  &56.76 &73.96 & 53.26 &71.34 \cr
    
    MatchingNet-LFT \cite{Hungfewshot} & 54.34  &69.92  & 58.41 &74.32  & 56.83 &71.62 \cr
    
    DPGN \cite{yang2020dpgn} & 54.83& 70.27& 58.95&74.56&57.18 &72.02 \cr 
    % \hline
    % FUNIT~\cite{liu2019few} &  &  &  &  &  & \cr

    DeepEMD \cite{zhang2020deepemd} & 54.15 &70.35  &59.12 &73.97  &58.01  &72.71 \cr
    % BSNet~\cite{li2020bsnet} &53.13 &71.56  &57.06 &72.83 & 56.86 &71.15 \cr
    GCNET \cite{Liu9343776} &53.73 &71.68  &57.61 &72.47 & 56.64 &71.53 \cr
    
    Delta-encoder \cite{schwartz2018delta} &53.19 &67.57 & 56.05& 72.84&56.38 & 71.29  \cr
    
    MatchingGAN \cite{hong2020matchinggan} & - & 70.94 &- & 74.09 & - &  70.89\cr

    F2GAN \cite{hong2020f2gan} &-  &72.31 &- &75.02  &-  &73.19 \cr
    LoFGAN \cite{gu2021lofgan}& - &73.01 &- &75.86 &- &73.43 \cr
    \hline
    DeltaGAN &$\textbf{56.8}$5 & $\textbf{75.71}$&$\textbf{61.23}$ & $\textbf{77.09}$&$\textbf{60.31}$ & $\textbf{74.59}$\cr
    % \Xhline{1.2pt}
    \bottomrule[0.8pt]
    \end{tabular}
    }
  \label{tab:performance_fewshot_classifier}
\end{table}

We compare our DeltaGAN with existing few-shot classification methods, including the representative methods MatchingNets \cite{vinyals2016matching}, RelationNets \cite{sung2018learning}, MAML \cite{finn2017model} as well as the state-of-the-art methods MTL \cite{sun2019meta},  MatchingNet-LFT \cite{Hungfewshot}, DPGN \cite{yang2020dpgn}, DeepEMD \cite{zhang2020deepemd}, and GCNET \cite{Liu9343776}.
% We compare our DeltaGAN with existing few-shot classification methods, including the representative methods MAML~\cite{finn2017model} as well as the state-of-the-art methods 
% % DN4~\cite{li2019revisiting}, 
% DPGN~\cite{yang2020dpgn}, and GCNET~\cite{Liu9343776}.
% % and BSNet~\cite{li2020bsnet}. 
For these baselines, no augmented images are added to the training set in each evaluation episode. Instead, the images from seen categories are used to train those few-shot classifiers by strictly following their original training procedure. 
% MAML \cite{finn2017model} and MTL~\cite{sun2019meta} models need to be finetuned based on the training set in each evaluation episode.
We also compare our DeltaGAN with few-shot image generation methods MatchingGAN and F2GAN as well as few-shot feature generation method Delta-encoder. We adopt the same augmentation strategy as our DeltaGAN in each evaluation episode. 
Besides, we compare our DeltaGAN with few-shot image translation method FUNIT~\cite{liu2019few} in Supplementary.
By taking $10$-way $1$-shot/$5$-shot as examples, we report the averaged accuracy over $10$ episodes on three datasets in Table~\ref{tab:performance_fewshot_classifier}.
Our method achieves the best performance on all datasets compared with few-shot classification and few-shot generation baselines, which demonstrates the high quality of generated images by our DeltaGAN.

\subsection{Ablation Studies}\label{sec:ablation}
% \label{sec:abalation}
We analyze the impact of each loss and alternative network designs on Animal Faces dataset in 1-shot setting. For each ablated method, FID, LPIPS, and the accuracy of $10$-way $1$-shot classification augmented with generated images are reported in Table~\ref{tab:network_design}. 
% More results of other ablated methods can be found in Supplementary.

\noindent\textbf{Loss terms: }In our method, we employ a reconstruction loss $\mathcal{L}_{1}$, a mode seeking loss $\mathcal{L}_{ms}$, a feature matching loss $\mathcal{L}_{fm}$, a classification loss $\mathcal{L}_{c}$, and an adversarial loss $\mathcal{L}^I_{adv}$.  
To investigate the impact of each loss term, we conduct ablation studies on Animal Faces dataset by removing each loss term from the final objective in Eqn.~\ref{optimization} separately. The results are summarized in Table~\ref{tab:network_design}, which shows that the diversity and fidelity of generated images are compromised when removing $\mathcal{L}_1$. By removing mode seeking loss $\mathcal{L}_{ms}$, we can see that all metrics become much worse, which implies the mode collapse issue after removing $\mathcal{L}_{ms}$. Another observation is that ablating $\mathcal{L}_{fm}$ leads to slight degradation of generated images. Removing $\mathcal{L}_{c}$ results in severe degradation of generated images, since the generated images may not belong to the category of conditional image. When $\mathcal{L}^I_{adv}$ is removed from the final objective, the worse quality of generated images indicates that typical adversarial loss can ensure the fidelity of generated images. 
% hyper-parameters (\emph{i.e.},$\lambda_1$, $\lambda_m$, $\lambda_s$,
% sample-specific delta space and conditional discriminator. 
% \noindent\textbf{Adversarial delta matching loss:}
To investigate the impact of our adversarial delta matching loss $\mathcal{L}^M_{adv}$ in Eqn.~\ref{eqn:adversarial_delta_loss}, 
We remove $\mathcal{L}^M_{adv}$ from the final objective in Eqn.~\ref{optimization}, which is referred to as ``w/o $\mathcal{L}^M_{adv}$'' in Table~\ref{tab:network_design}. We can see that the diversity and fidelity of generated images are compromised without $\mathcal{L}^M_{adv}$, because $\mathcal{L}^M_{adv}$ can bridge the gap between real delta and fake delta.

\setlength{\tabcolsep}{3pt}
\begin{table}[t]
  \caption{Ablation studies of our loss terms and alternative network designs on Animal Faces dataset} 
  \centering
   \resizebox{0.55\columnwidth}{!} {
  %\fontsize{8}{8}\selectfont
  \begin{tabular}{l|ccc}  
  \toprule[0.5pt]
     Setting& Accuracy(\%) $\uparrow$ & FID $\downarrow$  & LPIPS $\uparrow$ \cr
    \midrule
     w/o $\mathcal{L}_{1}$ & 58.68 & 100.21   &  0.4191   \cr
    
     w/o $\mathcal{L}_{ms}$ & 50.08 &121.74   &  0.2976  \cr
    
     w/o $\mathcal{L}_{fm}$ & 59.17 & 95.82   & 0.4324     \cr
     
     w/o $\mathcal{L}_{c}$ & 42.21 &  196.18  &  0.4119    \cr
     
     w/o $\mathcal{L}^I_{adv}$ & 52.18  &  139.46  &  0.3912 \cr
     
     w/o $\mathcal{L}^M_{adv}$ & 57.12 & 115.11    & 0.4153     \cr
    \midrule
     w/o real delta &53.03 & 128.69&0.3838 \cr
     \midrule
    % Prior delta &46.25 & 208.31&0.2431 \cr
    Global delta & 58.96 & 94.51   & 0.4311   \cr
    SC delta &56.11 &101.05 & 0.4162 \cr
    DC delta & 55.29 &105.91 &0.4021 \cr
    % simple $D_4$ &  & 104.56 & 0.4753       \cr
    \hline
     Simple $D_1$ & 54.53 & 129.17 & 0.3012       \cr
    Simple $D_2$ & 58.01 & 109.54 & 0.4401        \cr
    Simple $D_3$ & 59.51 & 94.12 &  0.4392       \cr
    Linear delta & 53.89 & 122.71 & 0.4091 \cr

    \midrule
    DeltaGAN & $\textbf{60.31}$ &$\textbf{89.81}$  & $\textbf{0.4418}$  \cr
    \bottomrule[0.8pt]
  \end{tabular}
  }
%   \vspace{0.1mm}
  \label{tab:network_design}
\end{table}

\noindent\textbf{Without real delta: }
To investigate the necessity of enforcing generated fake deltas to be close to real deltas, we cut off the links between real delta and fake delta by removing the reconstruction subnetwork and adversarial delta matching loss (\emph{i.e.}, removing $\{\mathcal{L}^M_{adv}, \mathcal{L}_{1}, \mathcal{L}_{fm}\}$), which is referred to as ``w/o real delta'' in Table~\ref{tab:network_design}. Compared with DeltaGAN, both diversity and realism are significantly degraded, because generation subnetwork fails to generate meaningful deltas without the guidance of reconstruction subnetwork and adversarial delta matching loss. 
Thus, we conclude that mode seeking loss needs to cooperate with our framework to produce realistic and diverse images.  
Another observation is that ``w/o $\mathcal{L}^M_{adv}$'' is better than ``w/o real delta'', which can be explained as follows. Even without using adversarial delta matching loss, since the reconstruction subnetwork and the generation subnetwork share the same $E_c$ and $G$, generated fake deltas have been implicitly pulled close to real deltas.

\noindent\textbf{Sample-specific delta: }
To corroborate the superiority of sample-specific delta, we directly use random vectors to generate deltas, which is referred to as ``Global delta'' in Table~\ref{tab:network_design}. It can be seen that our design of sample-specific deltas can benefit the quality of generated images. Besides, with our trained DeltaGAN model, we exchange sample-specific deltas within images from the same category (\emph{resp.,} across different categories) to generate new images, which is referred to as ``SC delta'' (\emph{resp.,} ``DC delta'') in Table~\ref{tab:network_design}. Compared with ``SC delta'' and ``DC delta'', our DeltaGAN achieves the best performance on all metrics, which verifies our assumption that delta is sample-specific and exchangeable use of deltas may lead to performance drop. We also visualize some examples generated by ``SC delta'' (\emph{resp.,} ``DC delta'') in Supplementary.

\noindent\textbf{Delta matching discriminator: }
In Section 3.2, we use conditional image, sample-specific delta, and output image as input triplet $\{\hat{D}_{I}(\bm{x}_1),\bm{\Delta}_{x_1}, \hat{D}_{I}(\bm{x}_2)\}$ for our delta matching discriminator $D_M$, which judges whether the conditional-output image pair matches the corresponding sample-specific delta. 
%To evaluate the effectiveness of our delta matching discriminator, we remove delta matching discriminator from our model,
% % we only use a traditional discriminator to distinguish real images from generated images without delta matching, 
% which is referred to as ``Simple $D_1$'' in Table~\ref{tab:network_design}. 
To evaluate the effectiveness and necessity of this input format, we explore different types of inputs for delta matching discriminator. As shown in Table~\ref{tab:network_design}, we use $\{ \bm{\Delta}_{x_1}\}$ (\emph{resp.}, $\{\hat{D}_{I}(\bm{x}_1),  \bm{\Delta}_{x_1}\}$, $\{\hat{D}_{I}(\bm{x}_2),  \bm{\Delta}_{x_1}\}$) as inputs of $D_M$, which is referred to as ``Simple $D_1$'' (\emph{resp.},``Simple $D_2$'',``Simple $D_3$''). We can see that ``Simple $D_1$'' is the worst, which demonstrates that only employing adversarial loss on delta does not work well. Besides,  both ``Simple $D_2$'' and ``Simple $D_3$'' are worse than our DeltaGAN, which demonstrates the effectiveness of matching conditional-output image pair with the corresponding sample-specific delta.

\noindent\textbf{Linear offset delta: }
To evaluate the effect of the learned non-linear ``delta'', we replace the non-linear ``delta''  with linear ``delta'', which is referred to as  ``Linear delta'' in Table~\ref{tab:network_design}. In the reconstruction subnetwork, we set $\bm{\Delta}^r_{x_1} = E_{\Delta} (\bm{x}_2) - E_{\Delta} (\bm{x}_1)$, and $\bm{\hat{x}}_2 = G(\bm{\Delta}^r_{x_1} +  E_{c} (\bm{x}_1)$), which means that we simply learn linear offset ``delta'' from same-class pairs of training data. In the generation subnetwork, we apply the generated fake  ``delta'' $\bm{\Delta}^f_{x_1}$ to conditional image $\bm{x}_1$ to generate new image $\bm{\tilde{x}}_2 = G(\bm{\Delta}^f_{x_1} +  E_{c} (\bm{x}_1))$. Based on Table~\ref{tab:network_design}, the FID gap between ``Linear delta'' and ``DeltaGAN'' indicates that complex transformations of intra-category pairs cannot be well captured by linear offset.

\section{Conclusion}
% In this paper, we have explored applying sample-specific deltas to a conditional image to generate new images by learning a reconstruction subnetwork and a generation subnetwork. 
% % Specifically, we have proposed a novel few-shot generation method DeltaGAN composed of a reconstruction subnetwork and a generation subnetwork, which are bridged by an adversarial delta matching loss.
% The experimental results on four datasets have shown that our DeltaGAN can substantially improve the quality and diversity of generated images.
% % compared with existing few-shot image generation methods.
% % The experimental results on six datasets have shown that our DeltaGAN can substantially improve the quality and diversity of generated images compared with existing few-shot image generation methods.
% % to extract sample-specific deltas from same-category pairs of seen categories and apply sampled random deltas from learnt delta space to unseen categories.
% % We have conducted extensive generation and classification experiments on five datasets to demonstrated the effectiveness of our method.

In this paper, we have explored applying sample-specific deltas to a conditional image to generate new images. Specifically, we have proposed a novel few-shot generation method DeltaGAN composed of a reconstruction subnetwork and a generation subnetwork, which are bridged by an adversarial delta matching loss.
The experimental results on six datasets have shown that our DeltaGAN can substantially improve the quality and diversity of generated images compared with existing few-shot image generation methods.

\section*{Acknowledgement}
The work is supported by Shanghai Municipal Science and Technology Key Project (Grant No. 20511100300), Shanghai Municipal Science and Technology Major Project, China (2021SHZDZX0102), and National Science Foundation of China (Grant No. 61902247).

\clearpage
% ---- Bibliography ----
%
% BibTeX users should specify bibliography style 'splncs04'.
% References will then be sorted and formatted in the correct style.
%
\bibliographystyle{splncs04}
\bibliography{egbib}
\end{document}

% \renewcommand\thelinenumber{\color[rgb]{0.2,0.5,0.8}\normalfont\sffamily\scriptsize\arabic{linenumber}\color[rgb]{0,0,0}}
% \renewcommand\makeLineNumber {\hss\thelinenumber\ \hspace{6mm} \rlap{\hskip\textwidth\ \hspace{6.5mm}\thelinenumber}}
% \linenumbers
\pagestyle{headings}
\mainmatter
\def\ECCVSubNumber{4032}  % Insert your submission number here

\title{Supplementary for DeltaGAN: Towards Diverse Few-shot Image Generation with Sample-Specific Delta} % Replace with your title

% INITIAL SUBMISSION 
\begin{comment}
\titlerunning{ECCV-22 4032 \ECCVSubNumber} 
\authorrunning{ECCV-22 4032 \ECCVSubNumber} 
\author{Anonymous ECCV submission}
\institute{4032 \ECCVSubNumber}
\end{comment}
%******************

% CAMERA READY SUBMISSION
%\begin{comment}
\titlerunning{DeltaGAN}
% If the paper title is too long for the running head, you can set
% an abbreviated paper title here
%
\author{Yan Hong\orcidlink{0000-0001-6401-0812}\index{Yan Hong} \and
Li Niu\thanks{Corresponding author.}  \orcidlink{0000-0003-1970-8634} \and
Jianfu Zhang\orcidlink{0000-0002-2673-5860} \and
Liqing Zhang\orcidlink{0000-0001-7597-8503}
}

\authorrunning{Y. Hong et al.}

\institute{MoE Key Lab of Artificial Intelligence, Shanghai Jiao Tong University, China \email{yanhong.sjtu@gmail.com, \{ustcnewly,c.sis\}@sjtu.edu.cn, zhang-lq@cs.sjtu.edu.cn}}
%\end{comment}
%******************

\maketitle

In this document, we provide additional material to support our main submission. In Section~\ref{sec:dataset_split}, we detail the split setting of datasets used in our paper. In Section~\ref{sec:architecture}, we describe the structure of our reconstruction subnetwork, generation subnetwork, and discriminators. In Section~\ref{sec:more_generation_visualization}, we visualize more images generated from our DeltaGAN. 
In Section~\ref{sec:low_data_classification}, we report the results of low-data classification augmented by generated images. 
In Section~\ref{sec:interpolation}, we show the interpolation results of our DeltaGAN on three datasets. In Section~\ref{sec:reconstruction}, we show some reconstructed images from our reconstruction subnetwork.
% In Section~\ref{sec:ablation_study_supple}, we provide results of ablated methods. 
In Section~\ref{sec:exchange_delta}, we show some images by exchanging deltas.
In Section~\ref{sec:few_shot_ability}, we compare our DeltaGAN with baselines in $K$-shot setting with different $K$.  In Section~\ref{sec:comparison_with_finetune_baseline}, we report comparison results between our method and other few-shot image generation methods relying on finetuning in the test phase.
In Section~\ref{sec:comparison_with_translation}, we compare our DeltaGAN with few-shot image translation method FUNIT. % In Section~\ref{sec:significant_test}, we conduct significance test for our DeltaGAN and baseline F2GAN.  
In Section~\ref{sec:failurecase}, we further analyze the limitation of our DeltaGAN.

\section{Datasets and Implementation Details}
\label{sec:dataset_split}
\textbf{Datasets}
We conduct experiments on six few-shot image datasets: EMNIST~\cite{cohen2017emnist},  VGGFace~\cite{cao2018vggface2}, Flowers~\cite{nilsback2008automated}, Animal Faces~\cite{deng2009imagenet}, NABirds~\cite{van2015building}, and Foods~\cite{kawano2014automatic}. Following the split setting of MatchingGAN~\cite{hong2020matchinggan} (\emph{resp.}, FUNIT~\cite{liu2019few}), we split VGGFace and  EMNIST  (\emph{resp.}, Animal Faces, Flowers, NABirds, and Foods) into seen categories and unseen categories. In detail, for VGGFace (\emph{resp.}, EMNIST) dataset, we randomly select 1802 (\emph{resp.}, 28) categories from all categories as seen training categories and select 96 (\emph{resp.},10) categories from remaining categories as unseen testing categories. On Flowers (\emph{resp.}, Animal Faces, NABirds, and Foods) dataset, a total of $102$ (\emph{resp.}, $149$, $555$, and $256$) categories are split into $85$ (\emph{resp.}, $119$, $444$, and $224$) seen categories and $17$ (\emph{resp.}, $30$, $111$, and $32$) unseen categories. In Table~\ref{tab:dataset_splite}, we summarize the number of seen/unseen categories and the number of seen/unseen images.

\noindent\textbf{Baselines} We compare our DeltaGAN with FIGR~\cite{clouatre2019figr}, DAWSON~\cite{liang2020dawson}, GMN~\cite{bartunov2018few}, DAGAN~\cite{antoniou2017data}, MatchingGAN~\cite{hong2020matchinggan}, F2GAN~\cite{hong2020f2gan}, and LoFGAN~\cite{gu2021lofgan}. For fair comparison, we conduct evaluation experiments for these methods using the same conditional images for each dataset. 
%The reported results of LoFGAN in our paper differ from those in the original paper, due to the usage of different conditional images in our paper compared to the original paper.
% The reported results of LoFGAN in our paper are different from original paper, which caused by using different conditional images compared with original paper.

% After having a few trials, we set $\lambda_1=10$, $\lambda_{fm} = 0.1$, and $\lambda_{ms} = 10$ by observing the quality of generated images during training. We adopt Adam optimizer with an initial learning rate of $1e-4$. The batch size is set to $16$ and our model is trained for $200$ epochs. 
\noindent\textbf{Implementation} We implement our model using the TensorFlow $1.13.1$ environment on Ubuntu $16.04$ LTS equipped by GEFORCE RTX $2080$ Ti GPU and Intel(R) Xeon(R) CPU $E5-2660$ v3 @ $2.60$GHz CPU. 
%To demonstrate the robustness of our method, we report the results of significance test by running our method and the best baseline for $10$ times with different random seeds (see Section~\ref{sec:significant_test}).

\setlength{\tabcolsep}{2pt}
\begin{table}[t]
  \caption{The splits of seen/unseen images (``img'') and categories (``cat'') on six datasets} 
  \centering
  %\fontsize{8}{8}\selectfont
   \resizebox{0.5\columnwidth}{!} {
  \begin{tabular}{l|rr|rr}  
    \toprule[0.8pt]
     \multirow{2}{*}{Dataset}& \multicolumn{2}{c|}{Seen} &\multicolumn{2}{c}{Unseen}\cr
      &\#img & \#cat   &\#img & \#cat \cr
     \midrule
     EMNIST~\cite{cohen2017emnist} &  78400 & 28 &  28000&10 \cr
     VGGFace~\cite{cao2018vggface2} &180200&1802 & 9600&96 \cr
     Flowers~\cite{nilsback2008automated} & 7121&85 & 1068&17 \cr
     Animal Faces~\cite{deng2009imagenet} & 96621&119 & 20863&30   \cr
     NABirds~\cite{van2015building} & 38306&444 &  10221&111 \cr
      Foods~\cite{kawano2014automatic} & 27471& 224 &3924 &  32 \cr
    \bottomrule[0.8pt]
  \end{tabular}
  }
%   \vspace{0.1mm}
  \label{tab:dataset_splite}
\end{table}
% \setlength{\tabcolsep}{10pt}
% \begin{table}[t]
%   \caption{The number of model parameters of MatchingGAN~\cite{hong2020matchinggan}, F2GAN~\cite{hong2020f2gan}, and our DeltaGAN.} 
%   \centering
%   %\fontsize{8}{8}\selectfont
%   \begin{tabular}{l|r|r}  
%     \toprule[0.8pt]
%      Setting& Training phase   & Testing phase \cr
%     \midrule
%     MatchingGAN~\cite{hong2020matchinggan} & 30.3M & 9.8M \cr
%     \midrule
%     F2GAN~\cite{hong2020f2gan} &31.1M  &9.9M \cr
%     \midrule
%     DeltaGAN &33.1M &  8.1M \cr
%     \bottomrule[0.8pt]
%   \end{tabular}
% %   \vspace{0.1mm}
%   \label{tab:model_parameters}
% \end{table}
\setlength{\tabcolsep}{2pt}
\begin{table}[t]
  \caption{{Comparison of the number of model parameters  (Million) and test time (second) among different few-shot image generation methods}
%   of MatchingGAN~\cite{hong2020matchinggan}, F2GAN~\cite{hong2020f2gan}, and our DeltaGAN.
  } 
  \centering
   \resizebox{0.7\columnwidth}{!} {
  %\fontsize{8}{8}\selectfont
   \begin{tabular}{l|cc|c}  
    \toprule[0.8pt]
    \multirow{2}{*}{Method} & 
    \multicolumn{2}{c|}{Model parameters} &  \multirow{2}{*}{Test time}  \cr
     & Training phase   & Testing phase & \cr
    %  \cmidrule(c){1-1} \cmidrule(c){2-3}
    %   \cmidrule(c){4} 
    \hline
     FIGR~\cite{clouatre2019figr}&23.9M &6.7M&0.0083s\cr
     GMN~\cite{bartunov2018few}&25.1M &18.5M&0.0324s\cr
     DAWSON~\cite{liang2020dawson}& 26.6M&7.1M&0.0091s\cr
     DAGAN~\cite{antoniou2017data}& 28.1M  & 8.7M&0.0104s\cr
    MatchingGAN~\cite{hong2020matchinggan} & 28.9M & 9.3M&0.0113s \cr
    F2GAN~\cite{hong2020f2gan} &29.6M  &9.5M&0.0121s \cr
    LoFGAN~\cite{gu2021lofgan} &39.2M &7.9M  &0.0149S \cr
    \hline
    DeltaGAN &31.5M &  7.7M& 0.0098s \cr
    \bottomrule[0.8pt]
  \end{tabular}
  }
%   \vspace{0.1mm}
  \label{tab:model_parameters}
\end{table}

\section{Details of Network Architecture}\label{sec:architecture}
\textbf{Generator} Our generator consists of a reconstruction subnetwork and a generation subnetwork. Our reconstruction subnetwork (\emph{resp.}, generation subnetwork) is constructed by $3$ encoders including  $E_{\Delta}$, $E_c$, and $E_r$ (\emph{resp.}, $E_f$) and $1$ decoder $G$. Encoder $E_{\Delta}$ has $5$ residual blocks~(ResBlks), which consists of $4$ encoder blocks and $1$ intermediate block. Each encoder block contains $3$ convolutional layers with leaky ReLU and batch normalization followed by one downsampling layer, while the intermediate block contains $3$ convolutional layers with leaky ReLU and batch normalization. The structure of encoder $E_c$ is the same as encoder $E_{\Delta}$ without parameters sharing. Encoder $E_r$ consists of two Conv-LRelu-BN blocks, in which each block contains $1$ convolutional layer with leaky ReLU and batch normalization. Encoder $E_f$ also has two Conv-LRelu-BN blocks. The decoder $G$ consists of $4$ residual blocks (ResBlks), in which each block contains $3$ convolutional layers with leaky ReLU and batch normalization followed by one upsampling layer. 
% The architecture of our generator is summarized in Table~\ref{tab:generator}. 

\noindent\textbf{Discriminator} Our discriminator $D_I$ is analogous to that in~\cite{liu2019few}, which consists of one convolutional layer followed by four groups of 
ResBlk. %residual blocks (ResBlks). 
Each group of ResBlk is as follows: ResBlk-$k$ $\rightarrow$ ResBlk-$k$ $\rightarrow$ AvePool$2$x$2$, where ResBlk-$k$ is a ReLU first residual block~\cite{mescheder2018training} with the number of channels $k$ set as $64$, $128$, $256$, $512$ in four groups. We use one fully connected (FC) layer with $1$ output following global average pooling (GAP) to obtain the discriminator score. Our discriminator $D_M$ is constructed by four FC layers following GAP. The classifier shares the feature extractor with the discriminator $D_I$ and only replaces the last FC layer with another FC layer with the number of outputs being the number of seen categories. 
% The architecture of our discriminators $D_I$ and $D_M$ is summarized in Table ~\ref{tab:discriminator}.

\setlength{\tabcolsep}{2pt}
\begin{table*}[t]
  \caption{Accuracy(\%) of different methods on two datasets in low-data setting. Among few-shot image generation methods, only DAGAN and our DeltaGAN are applicable in $1$-sample setting } 
  \centering
  %\fontsize{8}{8}\selectfont
  \resizebox{0.85\columnwidth}{!} {
  \begin{tabular}{l|rrrr|rrrr}
      % \Xhline{1.2pt}
      \toprule[0.8pt]
      \multirow{2}{*}{Method}& \multicolumn{4}{c|}{EMNIST} &\multicolumn{4}{c}{VGGFace}\cr
      &1 &5   & 10 & 15  &1 & 5  & 10 & 15 \cr
      \cmidrule(r){1-1} \cmidrule(r){2-5}  \cmidrule(r){6-9}
    Standard  &50.14& 83.64 & 88.64 & 91.14 & 5.08&8.82 & 20.29 & 39.12\cr
    Traditional  &52.82& 84.62  & 89.63 & 92.07& 8.87& 9.12 &22.83  & 41.63 \cr
    FIGR~\cite{clouatre2019figr}  &-& 85.91  & 90.08 & 92.18 &-& 6.12  &  18.84& 32.13 \cr
    GMN~\cite{bartunov2018few}   &-& 84.12  & 91.21 & 92.09 &-& 5.23  & 15.61 &35.48\cr
    DAWSON~\cite{liang2020dawson}  &-& 83.63 & 90.72 & 91.83 &-& 5.27 &16.92 &30.61 \cr
    DAGAN~\cite{antoniou2017data}   &57.84&87.45  & 94.18& 95.58 &13.27&19.23 &35.12 & 44.36\cr
    MatchingGAN~\cite{hong2020matchinggan}   &-& 91.75 & 95.91 &96.29  &-&21.12  &40.95 & 50.12\cr
    F2GAN~\cite{hong2020f2gan} &-& 93.18& 97.01 &97.82 &-&24.76&43.21 & 53.42\cr
    LoFGAN~\cite{gu2021lofgan}& -&93.56&97.35&97.94&-&24.56&43.89&54.02\cr
    \midrule
    DeltaGAN &$\textbf{84.56}$&$\textbf{96.02}$ &$\textbf{98.12}$ &$\textbf{98.87}$ &$\textbf{22.91}$& $\textbf{28.91}$&$\textbf{50.19}$&$\textbf{58.72}$ \cr
    \bottomrule[0.8pt]
    
  \end{tabular}
  }
%   \vspace{0.1mm}
  \label{tab:performance_vallia_classifier}
\end{table*}

\noindent\textbf{The number of model parameters}
We compare the number of model parameters of our DeltaGAN with MatchingGAN~\cite{hong2020matchinggan} and F2GAN~\cite{hong2020f2gan} in the training stage and testing stage, respectively. In the training stage, the model parameters of generator and discriminator are trainable to complete two-player adversarial learning with seen categories, while only generator is used to generate new images for each unseen category in the testing stage. In particular, our DeltaGAN only uses generation subnetwork to generate new images without reconstruction subnetwork. In Table~\ref{tab:model_parameters}, we can see that our DeltaGAN uses fewer model parameters to generate images of better quality compared with MatchingGAN and F2GAN in the testing stage.

% \input{sections/figures_tables/visualization_comparison_supple}

% \noindent\textbf{Comparison with Few-shot Image Generation Baselines}
% We visualize more images generated by DAGAN, F2GAN, and our DeltaGAN in $3$-shot setting in Figure~\ref{fig:visualization_compare_supple}.  From comparison, we can see that images generated by DAGAN are close to conditional images, since the generator of DAGAN tend to ignore the injected noise. For F2GAN, the diversity of F2GAN is limited, due to insufficient fusion of conditional images. In contrast, our DeltaGAN method can produce diverse images with higher quality, which indicates the superior design of our DeltaGAN.

% \begin{figure}[t]
% \begin{center}
% \includegraphics[scale=0.18]{./figures/combo_supp_v.png}
% \end{center}
% \caption{Images generated by our DeltaGAN in 1-shot setting on six datasets in the left subfigure
% (from top to bottom: EMNIST, VGGFace, and Flowers) and the right subfigure (from top to bottom: Foods, Animal Faces, and NABirds.)
% The conditional images are in the leftmost column in both subfigure}
% \label{fig:visualization_supple} 
% \end{figure}

\begin{figure}[tbp]
\begin{center}
\includegraphics[scale=0.23]{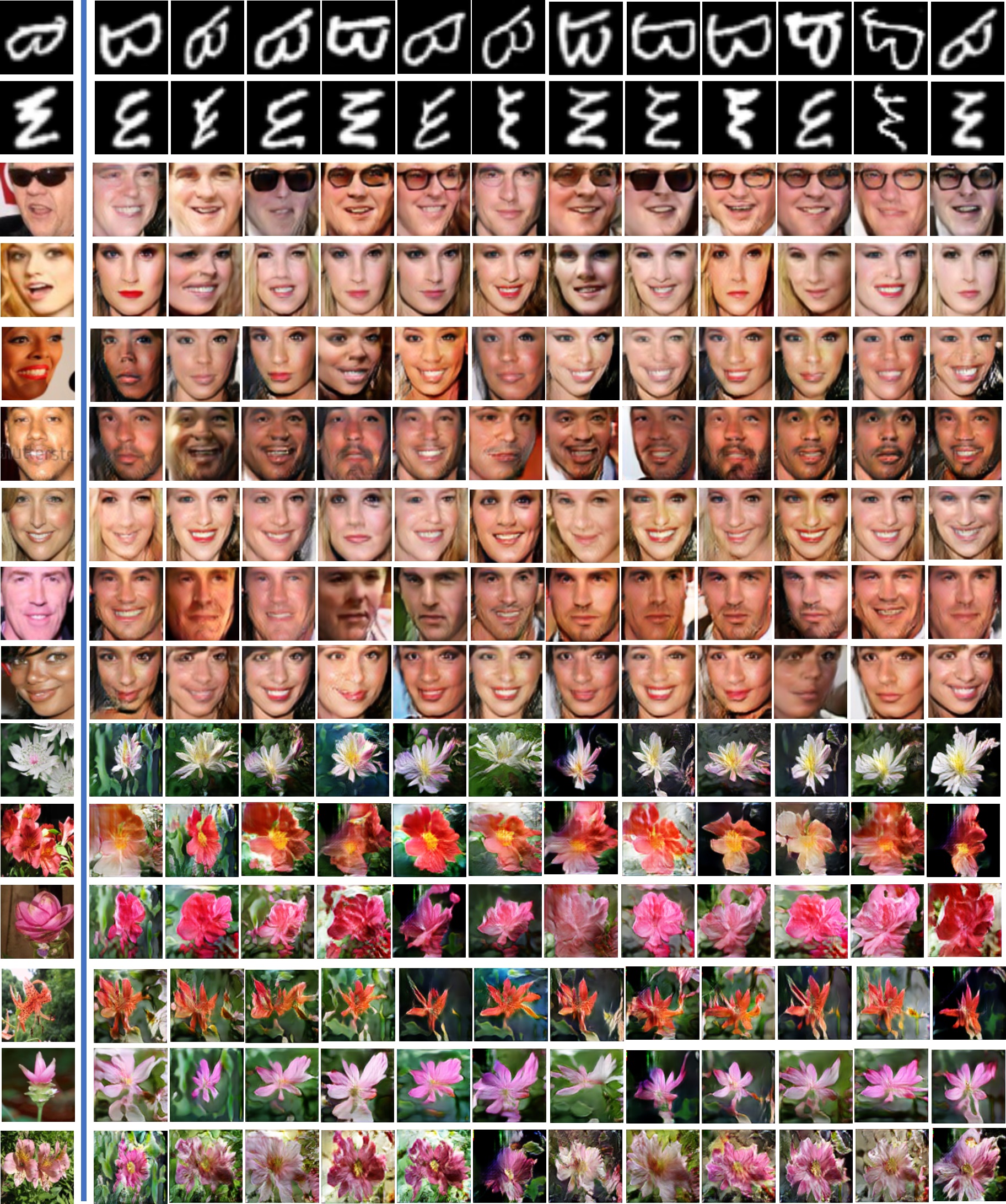}
\end{center}
\caption{Images generated by our DeltaGAN in 1-shot setting on three datasets.
From top to bottom: EMNIST, VGGFace, and Flowers.
The conditional images are in the leftmost column.}
\label{fig:visualization_supple} 
\end{figure}

\begin{figure}[tbp]
\begin{center}
\includegraphics[scale=0.23]{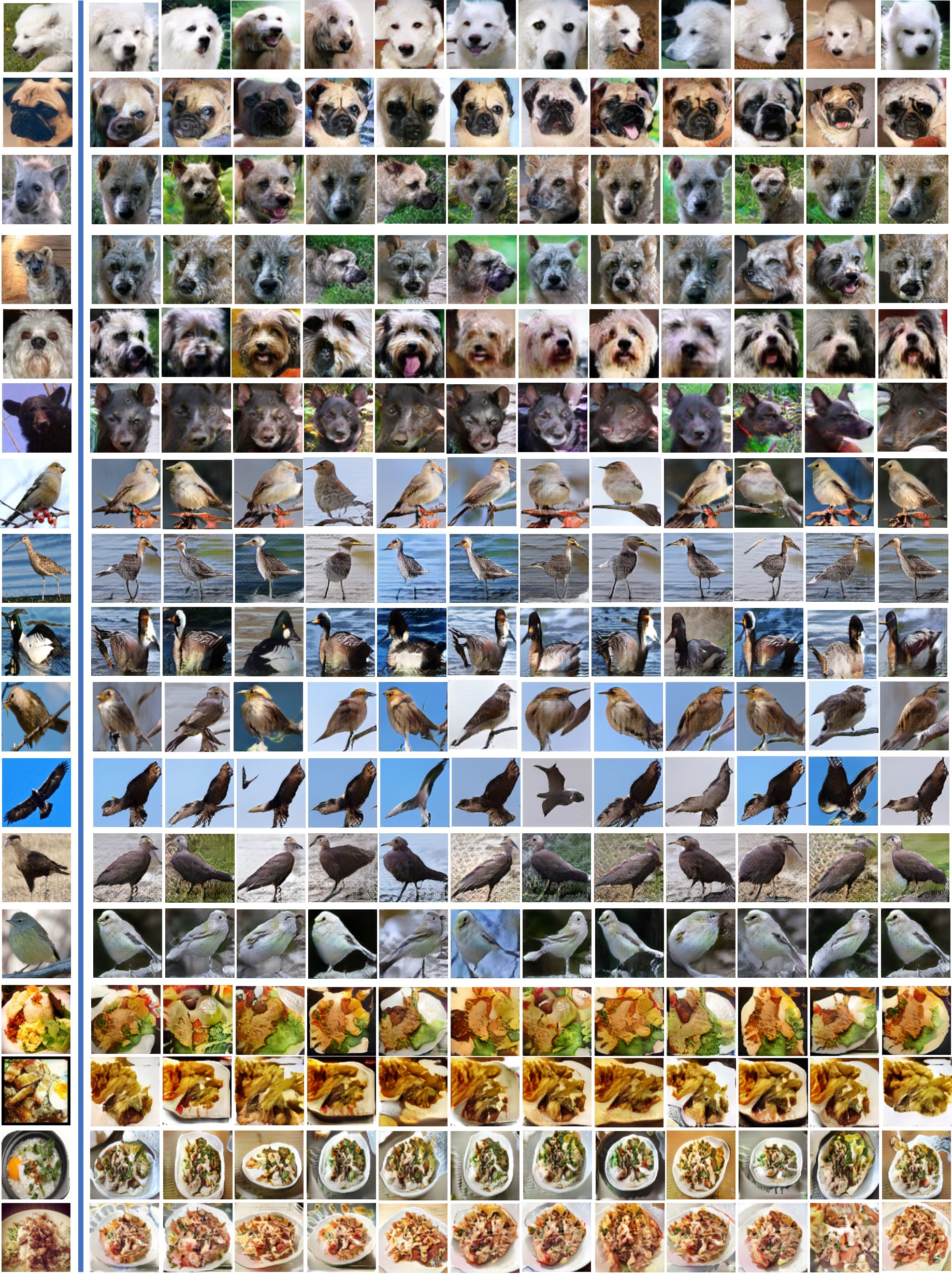}
\end{center}
\caption{Images generated by our DeltaGAN in 1-shot setting on three datasets. From top to bottom: Foods, Animal Faces, and NABirds.
The conditional images are in the leftmost column.}
\label{fig:visualization_supple_1} 
\end{figure}

\section{More Visualization Results}\label{sec:more_generation_visualization}
In this section, we visualize some example images generated by our DeltaGAN on EMNIST, VGGFace, Flowers, Animal Faces, NABirds, and Foods datasets in Fig.~\ref{fig:visualization_supple} and Fig.~\ref{fig:visualization_supple_1}.  On all datasets, our DeltaGAN can generally generate diverse and plausible images based on a single conditional image from unseen category.

\section{Low-data Classification}\label{sec:low_data_classification}
\label{sec:vallia}
To further evaluate the quality of our generated images, we conduct downstream classification tasks in low-data setting  by using generated images to augment unseen categories. Following F2GAN~\cite{hong2020f2gan}, for each unseen category, we randomly select a few (\emph{e.g.}, $K=1, 5, 10, 15$) training images and use the remaining images as test images, which is referred to as $K$-sample in Table~\ref{tab:performance_vallia_classifier}.
We initialize ResNet$18$~\cite{he2016deep} backbone based on seen categories,  then finetune the whole network with the training images of unseen categories, and finally apply the trained classifier to the test images of unseen categories. This setting is referred to as ``Standard'' in Table~\ref{tab:performance_vallia_classifier}.

\begin{figure}[t]
\begin{center}
\includegraphics[scale=0.5]{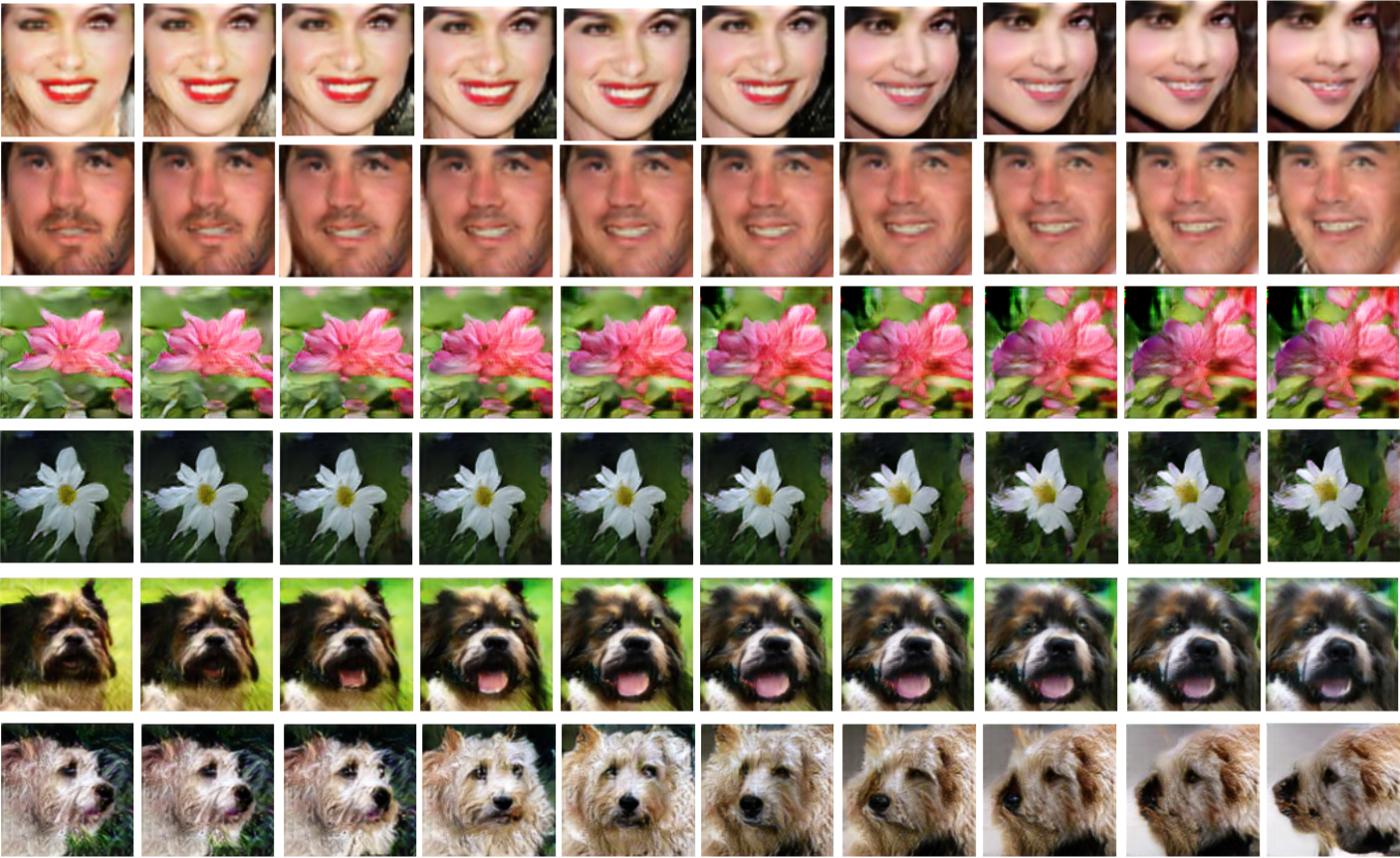}
\end{center}
\caption{Images generated by our DeltaGAN by interpolating random vectors between $\bm{z}_1$ and $\bm{z}_2$ on three datasets (from top to bottom: VGGFace, Flowers, and Animal Faces)}
\label{fig:interpolation} 
\end{figure}
\begin{figure}[t]
\begin{center}
\includegraphics[scale=0.48]{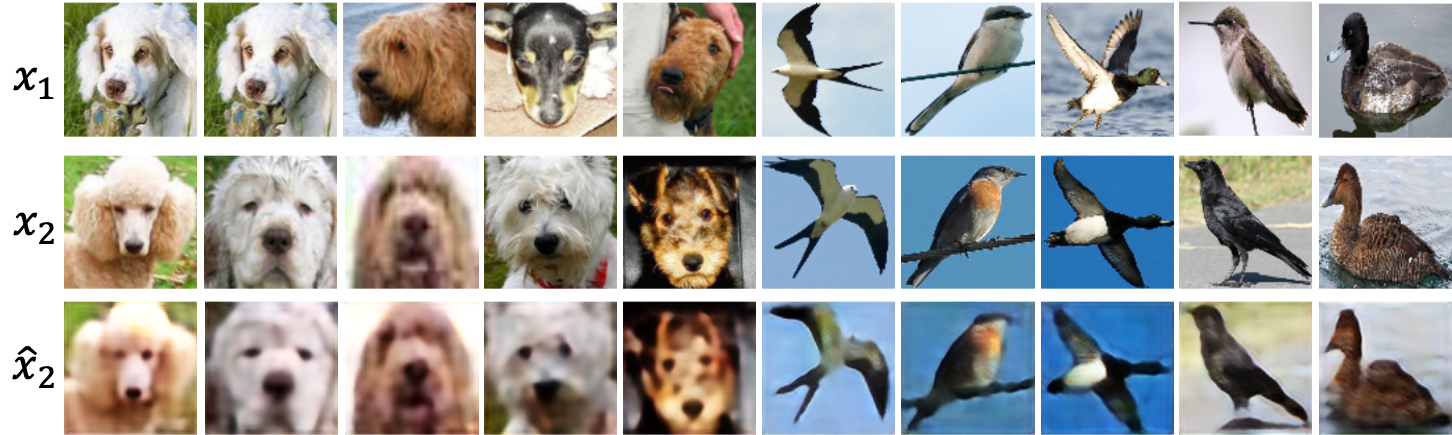}
\end{center}
\caption{Reconstruction results of our DeltaGAN on Animal Faces (left) and NABirds (right) datasets. The conditional images $\bm{x}_1$ are in the first row, the target images $\bm{x}_2$ are in the second row, and the reconstructed images $\hat{\bm{x}}_2$ are in the third row}
\label{fig:reconstructed} 
\end{figure}

Then, we augment unseen training images with new images generated by different few-shot image generation methods. For each unseen category, one method generates $512$ images by randomly sampling conditional images from the training set of this unseen category. Then, we augment the original training set of unseen categories with generated images, which are used to finetune the ResNet$18$ classifier. In addition, we compare with traditional data augmentation (\emph{e.g.}, flip, crop, color jittering), which also generates $512$ new images for each unseen category. The setting of traditional data augmentation is referred to as ``Traditional'' in Table~\ref{tab:performance_vallia_classifier}.
%Finally, the trained classifier is applied to the test set of unseen categories. 
The results of different methods are reported in Table~\ref{tab:performance_vallia_classifier}. We can see that our DeltaGAN achieves better results than traditional data augmentation methods as well as few-shot image generation baselines, which shows the effectiveness of using augmented images produced by our DeltaGAN for low-shot classification task.

\section{Delta Interpolation} \label{sec:interpolation}
To evaluate whether the delta space of DeltaGAN is densely populated, we perform linear interpolation based on two random vectors $\bm{z}_1$ and $\bm{z}_2$. In detail, we calculate the interpolated random vector $\bm{z}=a_1\bm{z}_1 + a_2\bm{z}_2$ ($a_1+a_2=1$) by gradually decreasing (\emph{resp.}, increasing) $a_1$ (\emph{resp.},  $a_2$) from $1$ (\emph{resp.}, $0$) to $0$ (\emph{resp.}, $1$) with step size $0.1$. We use one conditional image and interpolated $\bm{z}$ to generate sample-specific deltas, which are used to produce interpolation results in Fig.~\ref{fig:interpolation}. We can see that our DeltaGAN can generate diverse images with smooth transition between two random vectors $\bm{z}_1$ and $\bm{z}_2$, including the transition between different colors, shapes, and poses.

\section{Image Reconstruction Results}\label{sec:reconstruction}
In the training stage, our reconstruction network can reconstruct $\bm{x}_2$ based on $\bm{x}_1$ and $\bm{\Delta}_{x_1}^r$.
To demonstrate that the reconstruction ability of reconstruction subnetwork can be transferred from seen categories to unseen categories, we visualize the reconstructed unseen images on Animal Faces and NABirds datasets in Fig.~\ref{fig:reconstructed}. To be exact, we randomly sample same-category unseen image pairs $\{\bm{x}_1, \bm{x}_2\}$, which pass through our reconstruction network to yield $\hat{\bm{x}}_2$. From Fig.~\ref{fig:reconstructed}, we can see that the reconstructed images $\hat{\bm{x}}_2$ are quite close to the target images $\bm{x}_2$.

% \section{Ablation Study} \label{sec:ablation_study_supple}
% \noindent\textbf{Loss terms: }In our method, we employ a reconstruction loss $\mathcal{L}_{1}$, a mode seeking loss $\mathcal{L}_{ms}$, a feature matching loss $\mathcal{L}_{fm}$, a classification loss $\mathcal{L}_{c}$, and an adversarial loss $\mathcal{L}^I_{adv}$.  
% % Considering that our discriminator is based on MSGAN~\cite{mao2019mode}, below ablation study is conducted with $\mathcal{L}_{ms}$. 
% To investigate the impact of each loss term, we conduct ablation studies on Animal Faces dataset by removing each loss term from the final objective in Eqn. (11) in main paper separately. The results are summarized in Table~\ref{tab:network_design_supple}, which shows that the diversity and fidelity of generated images are compromised when removing $\mathcal{L}_1$. By removing mode seeking loss $\mathcal{L}_{ms}$, we can see that all metrics become much worse, which implies the mode collapse issue after removing $\mathcal{L}_{ms}$. Another observation is that ablating $\mathcal{L}_{fm}$ leads to slight degradation of generated images. Removing $\mathcal{L}_{c}$ results in severe degradation of generated images, since the generated images may not belong to the category of conditional image. When $\mathcal{L}^I_{adv}$ is removed from the final objective, the worse quality of generated images indicates that typical adversarial loss can ensure the fidelity of generated images. 

\begin{figure}[t]
\begin{center}
\includegraphics[scale=0.7]{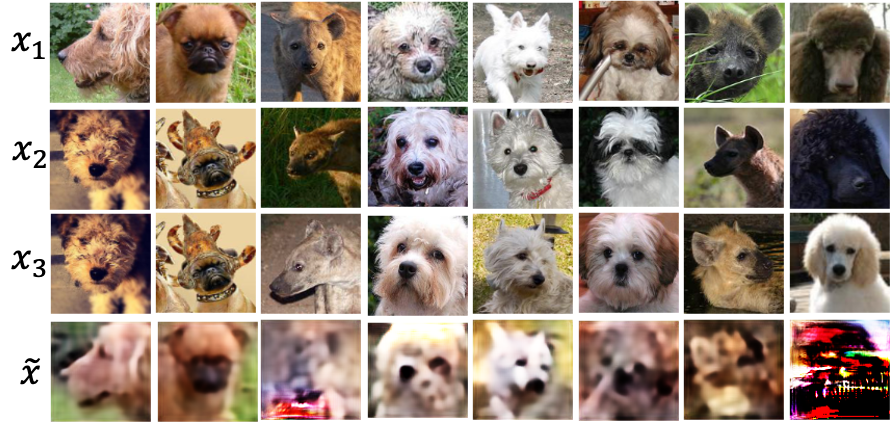}
\end{center}
\caption{Visualization results of exchanging delta in the reconstruction subnetwork on Animal Faces. From top to bottom: conditional image $\bm{x}_1$, a pair of $\bm{x}_2$ and $\bm{x}_3$ providing real delta, $\tilde{\bm{x}}$ generated based on $\bm{x}_1$ and the real delta}
\label{fig:deltavis} 
\end{figure}
\begin{figure}[t]
\begin{center}
\includegraphics[scale=0.55]{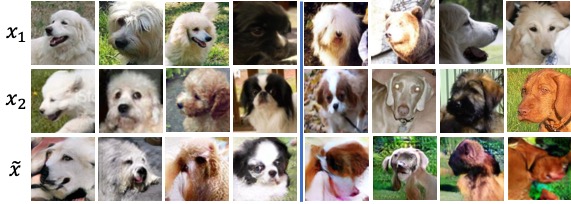}
\end{center}
\caption{Visualization results of exchanging delta in the generation subnetwork on Animal Faces. From top to bottom: conditional image $\bm{x}_1$, $\bm{x}_2$ providing the delta,  $\tilde{\bm{x}}$ generated based on $\bm{x}_1$ and the delta of $\bm{x}_2$}
\label{fig:scdc_delta} 
\end{figure}

\section{Visualization of Exchanging Delta}\label{sec:exchange_delta}

Note that our learnt delta is sample-specific delta. To check whether the delta is transferable across different images, we first show some generated images after exchanging delta in the reconstruction subnetwork.
As shown in Fig.~\ref{fig:deltavis},  we extract real delta $\bm{\Delta}_{23}^{r}$ from one pair of images $\{\bm{x}_2, \bm{x}_3\}$ from the same category as $\bm{x}_1$, after which $\bm{\Delta}_{23}^{r}$ is applied to the conditional image $\bm{x}_1$ to generate a new image $\tilde{\bm{x}}$. In column 1-2, we investigate a special case $\bm{x}_2 = \bm{x}_3$. In this case, the delta is vacuous and thus the generated image is close to the conditional image $\bm{x}_1$. Recall that the real delta $\bm{\Delta}_{23}^{r}$ between $\bm{x}_2$ and $\bm{x}_3$ contains the necessary information required to transform  $\bm{x}_2$ to  $\bm{x}_3$.
In column 3-6, we show some cases, in which the delta appearance information or delta pose information encoded in $\bm{\Delta}_{23}^{r}$ influences $\bm{x}_1$ to some degree. For example, in column 3, the fur color of $\bm{x}_3$ is lighter than $\bm{x}_2$, so the fur color of generated image $\tilde{\bm{x}}$ is also lighter than $\bm{x}_1$. In column 5, $\bm{x}_3$ turns face to the left compared with  $\bm{x}_2$, so $\tilde{\bm{x}}$ also turns face to the left compared with $\bm{x}_1$. However, the generated images are generally of low quality. In column 7-8, the generated images $\tilde{\bm{x}}$ are corrupted when there is huge difference between $\bm{x}_2$ and $\bm{x}_3$ in appearance and pose. 

Additionally, we show the visualization results of exchanging delta in the generation subnetwork in Fig.~\ref{fig:scdc_delta}, in which we apply the delta provided by $\bm{x}_2$ to the conditional image $\bm{x}_1$ to generate a new image $\tilde{\bm{x}}$. The left (\emph{resp.}, right) four columns correspond to ``SC delta'' (\emph{resp.}, ``DC delta'') in Table 3 in main paper, in which $\bm{x}_2$ is from the same (\emph{resp.}, different) category of $\bm{x}_1$.
We can see that exchanging delta usually leads to severely degraded quality of generated images compared with DeltaGAN. Another observation is that ``DC delta'' is more inclined to generated unreasonable images compared with ``SC delta''. These observations coincide with the quantitative results of ``DC delta'' and ``SC delta'' in Table 3 in main paper.

As shown in Fig.~\ref{fig:deltavis} and Fig.~\ref{fig:scdc_delta}, a plausible delta for one conditional image may be unsuitable for another conditional image, so we target at learning sample-specific delta. The learnt sample-specific delta has weak transferability across images and weaker transferability across categories. Hence, it is not suggested to generate new images by exchangeably applying delta. However, the ability of generating effective sample-specific delta is transferable from seen categories to unseen categories, so our DeltaGAN is effective in producing new realistic images based on a conditional unseen image.

\begin{figure}[t]
\begin{center}
\includegraphics[scale=0.12]{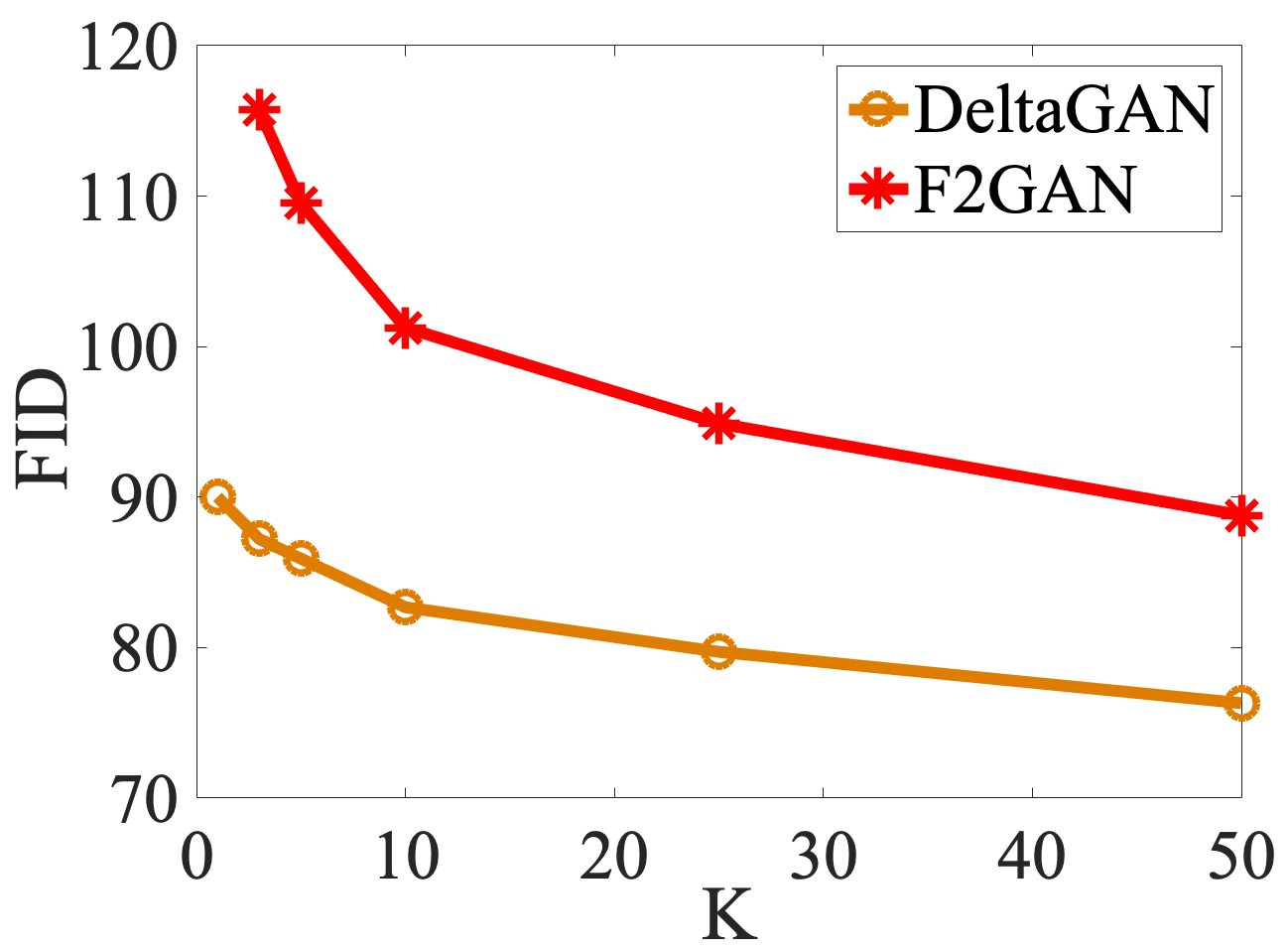}
\includegraphics[scale=0.12]{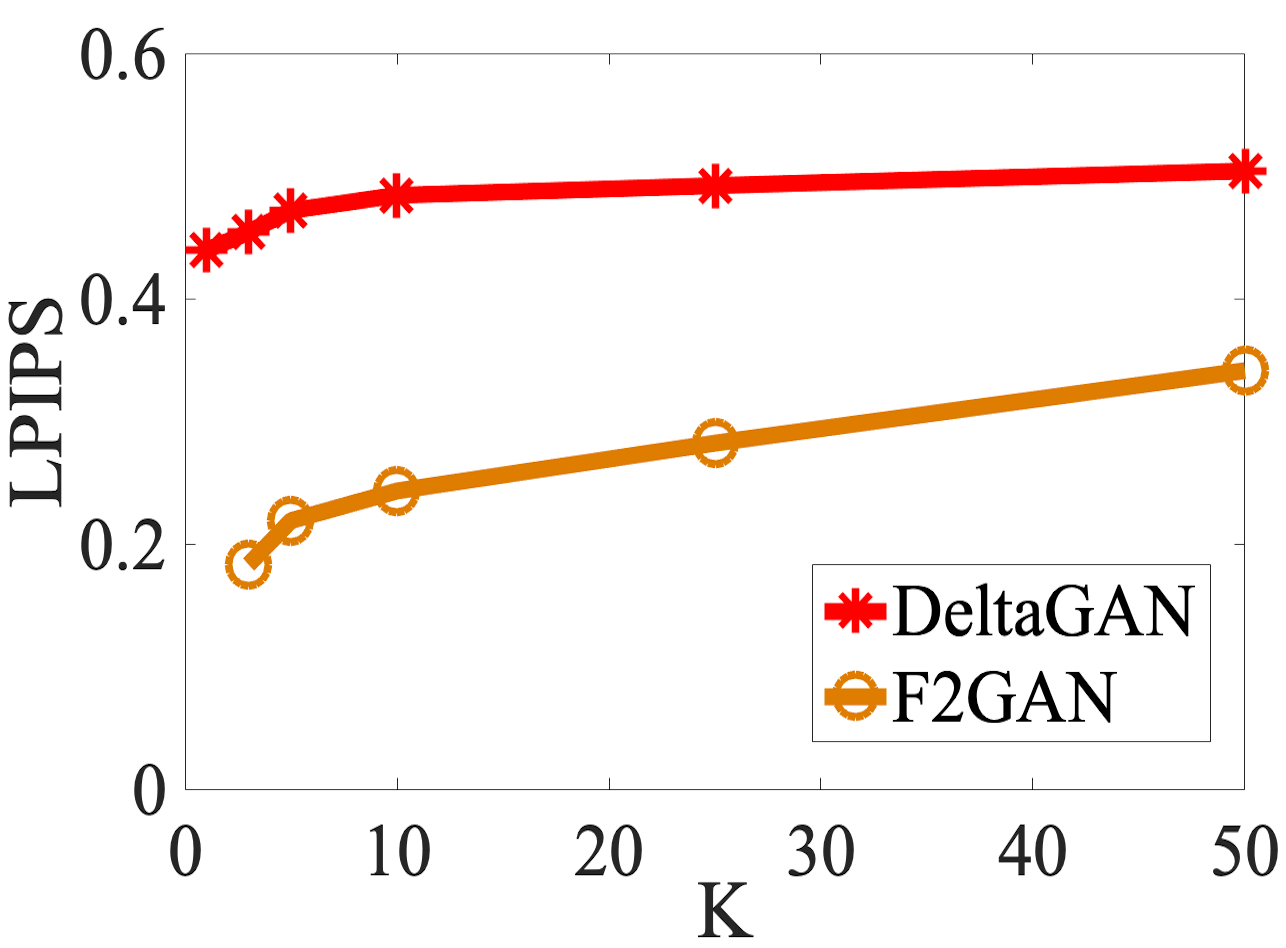}
\end{center}
\caption{FID and LPIPS comparison between F2GAN and our DeltaGAN with different numbers ($K$) of conditional images on Animal Faces} 
\label{fig:metrcis_curve} 
\end{figure}

\section{Few-shot Generation Ability}\label{sec:few_shot_ability}
Here, we repeat the experiments in Section 4.1 in the main paper except tuning $K$ in a wide range. Recall that $K$ in $K$-shot setting means that $K$ real images are provided for each unseen category. We use our DeltaGAN and competitive baseline F2GAN to generate 128 new images for each  unseen  category in $K$-shot setting. We also adopt the same quantitative evaluation metrics (FID and LPIPS) to measure image quality and diversity as in Section 4.1 in the main paper. We plot the FID curve and LPIPS curve of two methods with increasing $K$ in Fig.~\ref{fig:metrcis_curve}. It can be seen that our DeltaGAN outperforms F2GAN by a large margin with all values of $K$, especially when $K$ is very small. These results demonstrate that our DeltaGAN can generate abundant diverse and realistic images even if only a few (\emph{e.g.}, 10) real images are provided. 

\setlength{\tabcolsep}{4pt}
\begin{table}[t]
  \centering
  \resizebox{0.35\columnwidth}{!} {
%   \fontsize{8}{8}\selectfont
  \begin{tabular}{l|c|c}  
    \hline
     Setting&  FID $\downarrow$  & LPIPS $\uparrow$ \cr
    \hline
     \cite{ojha2021few} &109.89 & 0.2879 \cr
    \hline
    Ours & \textbf{109.78} & \textbf{0.3912}   \cr
    \hline
  \end{tabular}
  }
\caption{FID ($\downarrow$) and LPIPS ($\uparrow$) of images generated by \cite{ojha2021few} and our method on Flowers dataset} 
% \caption{Comparison between (Ojha et al. 2021) and our method on Flowers dataset.} 
\label{tab:comparison_metric_finetune}
\end{table}
\section{Comparison with Other Few-shot Image Generation} \label{sec:comparison_with_finetune_baseline}
As discussed in Section 2 in the main paper, our setting is quite different from recent works \cite{ojha2021few,li2020few,robb2020few,WangGBHK020}. Our method can achieve instant adaptation to multiple unseen categories without finetuning, while the abovementioned methods need to finetune the trained model for each unseen category, which is very resource-consuming and time-consuming. We compare the performance of \cite{ojha2021few} with our method on Flowers dataset.  For \cite{ojha2021few}, we train a source model on all seen images during training. At test stage, we finetune the source model on each selected unseen category with one image (1-shot setting in Section 4.1 in main paper) and produce $128$ images by sampling random vectors for evaluation.
We report the results in Table{~\ref{tab:comparison_metric_finetune}}. We can see that our method slightly outperforms \cite{ojha2021few} and produces more diverse and realistic images. However, \cite{ojha2021few} requires finetuning for each unseen category, while our model can be instantly adapted to unseen categories without finetuning.

\section{Comparison with Few-shot Image Translation}\label{sec:comparison_with_translation}
\begin{figure}[t]
\begin{center}
\includegraphics[scale=0.5]{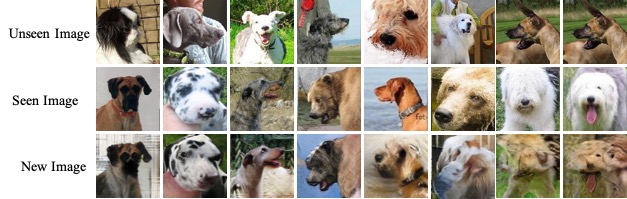}
\end{center}
\caption{Images generated by FUNIT~\cite{liu2019few} in 1-shot setting on Animal Faces. Unseen (\emph{resp.}, seen) images are shown in the first (\emph{resp.}, second) row. In each column, the new image is generated by combining the class code of unseen image and the content code of seen image} 
\label{fig:funit_1} 
\end{figure}

Recently, few-shot image translation methods like FUNIT~\cite{liu2019few} have been proposed to translate seen images to unseen categories, which can also generate new images for unseen categories given a few images. However, the motivations of few-shot image generation and few-shot image translation are considerably different. Specifically, the former can generate new unseen images without touching seen images, while the latter relies on seen images to generate new unseen images. In particular, FUNIT disentangles the latent representation of an image into category-relevant representation~(\emph{i.e.}, class code) and category-irrelevant representation~(\emph{i.e.}, content code), in which appearance belongs to class code and pose belongs to content code~\cite{liu2019few}. In the testing stage, given a few images from one unseen category, FUNIT generates new images for this unseen category by combining the content codes of seen images with the class codes of these unseen images. However, in reality, the disentanglement in FUNIT is not perfect and the content code may also contain appearance information. So when translating seen images to unseen categories, the appearance information of seen images may be leaked to the generated new images. To corroborate this point, we visualize some example images generated by FUNIT. As shown in Fig.~\ref{fig:funit_1}, in column 1-4, the generated new images contain the appearance of seen images, which is against our expectation that the generated new images should be from unseen categories. In column 5-8, the generated images are even corrupted, probably due to incompatible content codes and class codes.

\setlength{\tabcolsep}{2pt}
\begin{table}[t]
  \caption{Accuracy(\%) of different methods on Animal Faces in few-shot classification setting. Note that MatchingGAN, F2GAN, and LoFGAN are not applicable in 1-shot setting} 
  \centering
    \resizebox{0.6\columnwidth}{!} {
  %\fontsize{8}{8}\selectfont
  \begin{tabular}{l|rr}
      % \Xhline{1.2pt} 
      \toprule[0.8pt]
    %   \hline
      Method & 10-way 1-shot &10-way 5-shot\cr
    %   \cmidrule(r){2-3}  
    \midrule
    DPGN~\cite{yang2020dpgn} &57.18 &72.02 \cr 
    DeepEMD~\cite{zhang2020deepemd} &58.01  &72.71 \cr
    % \hline
    % FUNIT~\cite{liu2019few} &  &  &  &  &  & \cr
    MatchingGAN~\cite{hong2020matchinggan} & - & 70.89\cr
    F2GAN~\cite{hong2020f2gan} & - & 73.19 \cr
    LoFGAN~\cite{gu2021lofgan} & - &73.43 \cr
    \midrule
    FUNIT-1 & 56.61 &69.12  \cr
    FUNIT-2 & 53.38 &67.87  \cr
    \midrule
    DeltaGAN & $\textbf{60.31}$  & $\textbf{74.59}$ \cr
    % \Xhline{1.2pt}
    \bottomrule[0.8pt]
    \end{tabular}
    }
%   \vspace{0.1mm}
  \label{tab:performance_fewshot_funit}
\end{table}

% \setlength{\tabcolsep}{8pt}
% \begin{table}[t]
%   \caption{Accuracy(\%) of different methods on Animal Faces in few-shot classification setting. Note that MatchingGAN and F2GAN are not applicable in 1-shot setting.} 
%   \centering
%   %\fontsize{8}{8}\selectfont
%   \begin{tabular}{l|rr}
%       % \Xhline{1.2pt} 
%       \toprule[0.8pt]
%     %   \hline
%       Method & 10-way 1-shot &10-way 5-shot\cr
%     %   \cmidrule(r){2-3}  
%     \hline
%     MatchingNets~\cite{vinyals2016matching}   &36.54  &50.12 \cr

%     MAML~\cite{finn2017model} & 35.98  &49.89 \cr

%     RelationNets~\cite{sung2018learning} &45.32  & 58.12 \cr

%     MTL~\cite{sun2019meta} & 52.54 &70.91 \cr

%     DN4~\cite{li2019revisiting} & 53.26 &71.34 \cr
    
%     MatchingNet-LFT~\cite{Hungfewshot}  &56.84  &71.62 \cr
%     DPGN~\cite{yang2020dpgn} &57.18 & 72.02 \cr
%     Delta-encoder\cite{schwartz2018delta} & 56.38 & 71.29 \cr
%     DeepEMD~\cite{zhang2020deepemd}& \cr
%     BSNet~\cite{li2020bsnet} & \cr
%     % \hline
%     % FUNIT~\cite{liu2019few} &  &  &  &  &  & \cr
%     MatchingGAN~\cite{hong2020matchinggan} & - & 70.89\cr
%     F2GAN~\cite{hong2020f2gan} & - & 73.19 \cr
%     \hline
%     FUNIT-1 & 56.61 &69.12  \cr
%     FUNIT-2 & 53.38 &67.87  \cr
%     \hline
%     DeltaGAN & $\textbf{60.31}$  & $\textbf{74.59}$ \cr
%     % \Xhline{1.2pt}
%     \bottomrule[0.8pt]
%     \end{tabular}
% %   \vspace{0.1mm}
%   \label{tab:performance_fewshot_funit}
% \end{table}

%Thus, FUNIT fails to generate diverse images with only a few conditional images.
%Technically, FUNIT disentangles a given image into category-relevant factors~(\emph{i.e.}, class code) and category-irrelevant factors~(\emph{i.e.}, content code). Given a few images from an unseen category, FUNIT can generate more images for this unseen category by combining the content codes of abundant seen images with the class codes of a few unseen images.
% However, in this way, the generated images only have category-irrelevant diversity, while the category-relevant diversity in terms of category-specific properties is still limited by the small number of provided unseen images.
We also compare our DeltaGAN with FUNIT quantitatively for few-shot classification. By using the released model of FUNIT~\cite{liu2019few} trained on Animal Faces~\cite{deng2009imagenet}, we combine the content codes of seen images and the class codes of unseen images to produce $512$ new images for each unseen category. Then, the generated images are used to facilitate few-shot classification, which is recorded as ``FUNIT-1'' in Table~\ref{tab:performance_fewshot_funit}. However, ``FUNIT-1'' leverages seen images, which are not used in our DeltaGAN when generating new unseen images. For fair comparison, we also exchange content codes within the images from the same unseen category to produce new images, which is recorded as ``FUNIT-2'' in Table~\ref{tab:performance_fewshot_funit}. In this case, 
%In this setting, the content images and category images are both from the same unseen category, which can ensure that {\color{blue} both category code and content code of} the generated images belong to the same unseen category. However, the number of generated images is limited 
we can only generate $(C-1) \times C$ new images for each unseen category in $N$-way $C$-shot setting. 
Based on Table~\ref{tab:performance_fewshot_funit}, we observe that ``FUNIT-2'' is much worse than ``FUNIT-1'', because ``FUNIT-1'' resorts to a large number of extra seen images to generate more unseen images than ``FUNIT-2''. We also observe that ``FUNIT-1'' underperforms some few-shot classification methods and some few-shot image generation methods (\emph{e.g.}, F2GAN, DeltaGAN), which may be attributed to the appearance information leakage or image corruption as shown in Fig.~\ref{fig:funit_1}.
%which can be explained as follows. FUNIT only increases category-irrelevant diversity by borrowing the content codes from seen images, but does not bring in category-relevant diversity, which is intrinsically different from few-shot image generation methods. 
%The possible explanation is that the generated images from ``FUNIT-1'' fail to ensure the category information of class images, while ``FUNIT-2'' can only generate a limited number of new images. 
%We also show more example images generated by FUNIT in Figure~\ref{fig:funit}, it is seen that some generated images from FUNIT are unreasonable and of low-quality.

\setlength{\tabcolsep}{4pt}
\begin{table}[t]
  \caption{Significance test  between DeltaGAN and F2GAN on Animal Faces dataset in $3$-shot setting} 
    
  \centering
  \resizebox{0.7\columnwidth}{!} {
  %\fontsize{8}{8}\selectfont
  \begin{tabular}{lrrr}  
    \toprule[0.8pt]
     Setting& Accuracy(\%) $\uparrow$  & FID $\downarrow$  & LPIPS $\uparrow$ \cr
    \hline
    F2GAN & 75.65±0.32& 117.12±0.29 &0.1903±0.17 \cr
    \hline
    DeltaGAN & $\textbf{77.13±0.21}$ &$\textbf{87.12±0.06}$  & $\textbf{0.4661±0.11}$   \cr
    \bottomrule[0.8pt]
  \end{tabular}
  }
%   \vspace{0.1mm}
  \label{tab:significant_test}
\end{table}
\begin{figure}[t]
\begin{center}
\includegraphics[scale=0.24]{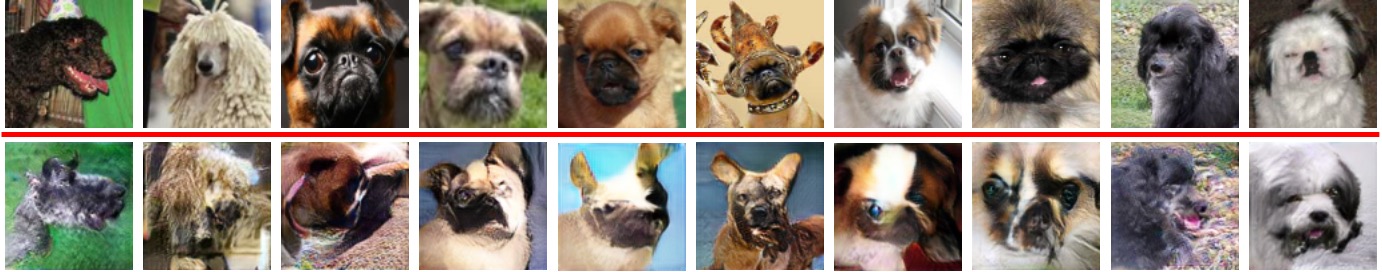}
\end{center}
\caption{Failure cases of our DeltaGAN on Animal Faces dataset. The conditional images are shown in the top row and the generated images are shown in the bottom row} 
\label{fig:failure_case} 
\end{figure}

%\section{Significance Test}\label{sec:significant_test}
%By taking Animal Faces dataset as an example, we run our DeltaGAN and the best baseline F2GAN for $10$ times with different random seeds.  The quality of generated images from our DeltaGAN and F2GAN is evaluated from two aspects. On one hand,  FID and LPIPS of generated images are computed as in Section 4.1 in main paper. Considering that the best baseline F2GAN is not applicable in $1$-shot learning, we conduct experiments in $3$-shot setting. On the other hand, we report the accuracy of few-shot classification ($10$-way $5$-shot) augmented with generated images as in Section 4.2 in the main paper. By conducting experiments $10$ times with different random seeds, we report the mean value (Mean) and standard deviation value (Std) in the form of Mean±Std for each evaluation metric and each method in Table~\ref{tab:significant_test}. Furthermore, we perform significance test at the significance level $0.05$ to verify that DeltaGAN outperforms the baseline F2GAN. The p value \emph{w.r.t.} few-shot classification (\emph{resp.}, FID and LPIPS) is $1.21e-5$ (resp., $2.34e-4$ and $3.89e-6$) and far below $0.05$, which demonstrates that the superiority of DeltaGAN is statistically significant.

\section{Limitation}

% \section{Failure Case} \label{sec:failurecase}
% We show some failure cases of our DeltaGAN on Animal Faces dataset in Fig.~\ref{fig:failure_case}. Due to the complexity of transformations between intra-category pairs, when applying the generated deltas to the conditional images (top row), some deltas may generate distorted images (bottom row).

\noindent\textbf{Failure cases:}\label{sec:failurecase}
Although our model achieves promising results both qualitatively and quantitatively on six datasets, some generated deltas are applied to the conditional images to produce distorted images due to the complexity of transformations between intra-category pairs. We show some failure cases of our DeltaGAN on Animal Faces dataset in Fig.~\ref{fig:failure_case}.

\begin{figure*}[t]
\begin{center}
\includegraphics[scale=0.1]{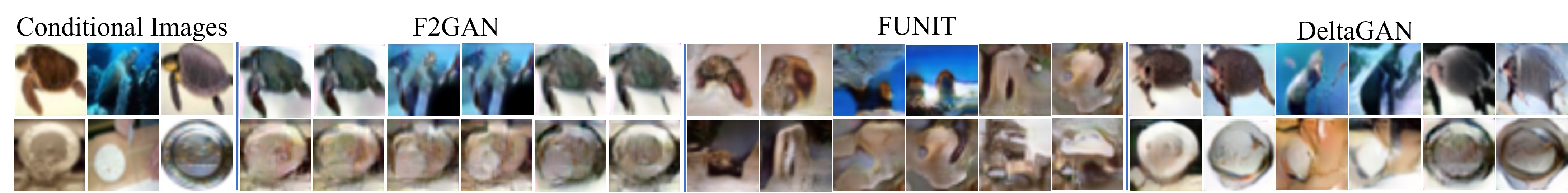}
\end{center}
\caption{Images generated by F2GAN, FUNIT and our DeltaGAN in 3-shot setting on CIFAR-100 dataset. The conditional images are in the left three column. The first row is category ``turtle", the second row is category ``plate".} 
\label{fig:cifar-100} 
\end{figure*}
\noindent\textbf{Generation ability on coarse-grained datasets:} Our method can transfer knowledge learned from seen categories to unseen categories to produce compelling results for unseen categories on fine-grained datasets. However, like other competitive few-shot image generation method F2GAN~\cite{hong2020f2gan} and few-shot image translation method FUNIT~\cite{liu2019few}, our method cannot achieve satisfactory results on coarse-grained datasets, such as CIFAR-100~\cite{krizhevsky2009learning} with large inter-category variance. We randomly divide a total of $100$ categories into $80$ seen training categories and $20$ unseen testing categories to conduct experiments for F2GAN, FUNIT, and our DeltaGAN. Similar to Section 4.1 in main paper, we  visualize some example images generated by different methods in 3-shot setting in Fig.~\ref{fig:cifar-100}. For FUNIT in 3-shot setting, we randomly select two seen images as content images to combine with each conditional image to produce new images.

We can observe that the structures of images generated by F2GAN are similar to conditional images. The generated images are vague and lacking local details. For FUNIT, the generated images do not have clear shape or overall structure. In contrast, the images produced by our DeltaGAN are relatively more diverse with clearer shape. We have to acknowledge that the quality of images generated by all three methods is poor, because it is difficult to achieve instant adaptation on coarse-grained datasets with large inter-category variance. 

\begin{figure}[t]
\begin{center}
\includegraphics[scale=0.45]{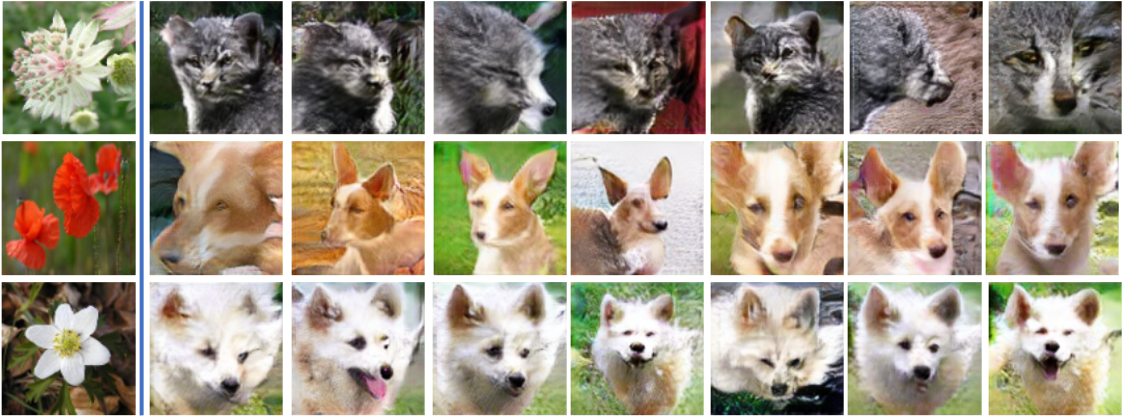}
\end{center}
\caption{Failure cases of our DeltaGAN trained on Animal Faces dataset and tested on Flowers dataset.The conditional images are in the leftmost column.} 
\label{fig:failure_case_adaptation} 
\end{figure}

\noindent\textbf{Adaptation ability between different datasets:} We also explore the adaptation ability of our DeltaGAN on Animal Faces dataset and Flowers dataset. In detail, we train DeltaGAN on Animals dataset, and test on Flowers dataset. Similar to Section 4.1 in main paper, we show some example images in Fig.~\ref{fig:failure_case_adaptation}. We can see that the generated images belong to the different animal face categories, although the given conditional images are from Flowers dataset.
The distributions of sample-specific deltas on different datasets are considerably different, so  DeltaGAN trained on Animal Faces dataset fails in capturing sample-specific deltas for Flowers dataset.
\bibliographystyle{splncs04}
\bibliography{egbib}